\documentclass[fleqn,10pt,top=2cm, bottom=2cm, left=2.5cm, right=3cm]{wlscirep}
\usepackage[utf8]{inputenc}
\usepackage[T1]{fontenc}
\usepackage{color}
\usepackage{multirow}
\usepackage[most]{tcolorbox}

\usepackage{amsthm,amssymb,amsmath,graphicx}
\usepackage{bm}
\usepackage{color}
\usepackage{amsmath}
\usepackage{nccmath} 
\usepackage{hyperref}

\usepackage{tabto}
\TabPositions{1.0cm, 3.0cm, 5.0cm, 7.0cm}

\usepackage[caption = false]{subfig}

\newtcbtheorem{Summary}{}{enhanced,drop shadow={black!50!white},
  coltitle=black,
  top=0.3in,
  attach boxed title to top left=
  {xshift=1.5em,yshift=-\tcboxedtitleheight/2},
  boxed title style={size=small,colback=pink}
}{summary}

\newtcbtheorem{Question}{}{enhanced,drop shadow={black!50!white},
  coltitle=black,
  top=0.3in,
  attach boxed title to top left=
  {xshift=1.5em,yshift=-\tcboxedtitleheight/2},
  boxed title style={size=small,colback=yellow}
}{Question}

\newtcbtheorem{Answer}{}{enhanced,drop shadow={black!50!white},
  coltitle=black,
  top=0.3in,
  attach boxed title to top left=
  {xshift=1.5em,yshift=-\tcboxedtitleheight/2},
  boxed title style={size=small,colback=green}
}{Answer}

\title{BioPars: A Pretrained Biomedical Large Language Model for Persian Biomedical Text Mining}

\author[1]{Baqer M. Merzah}
\author[2]{Tania Taami}
\author[3]{Salman Asoudeh}
\author[4]{Saeed Mirzaee}
\author[5]{Amir reza Hossein pour}
\author[6, *]{Amir Ali Bengari}

\affil[1]{Department of Computer Science, Faculty of Education, University of Kufa, Najaf, Iraq}

\affil[2]{Department of Computer Science, Florida State University, Tallahassee, USA}

\affil[3]{Department of Computer Engineering, Velayat University, Iranshahr, Iran}

\affil[4]{Department of Mechanical Engineering, Amirkabir University of Technology, Tehran, Iran}

\affil[5]{Department of Computer Engineering, Ferdowsi University of Mashhad, Mashhad, Iran}

\affil[6]{Department of Industrial and Technology Management, University of Tehran, Tehran, Iran}

\affil[*]{amirali.bengari@ut.ac.ir}

\affil[+]{these authors contributed equally to this work}

\begin{abstract}
Large Language Models (LLMs) have recently gained attention in the life sciences due to their capacity to model, extract, and apply complex biological information. Beyond their classical use as chatbots, these systems are increasingly used for complex analysis and problem-solving in specialized fields, including bioinformatics. First, we introduce BIOPARS-BENCH, a dataset from over 10,000 scientific articles, textbooks, and medical websites. BioParsQA was also introduced to evaluate the proposed model, which consists of 5,231 Persian medical questions and answers. This study then introduces BioPars, a simple but accurate measure designed to assess LLMs for three main abilities: acquiring subject-specific knowledge, interpreting and synthesizing such knowledge, and demonstrating proper evidence. Comparing ChatGPT, Llama, and Galactica, our study highlights their ability to remember and retrieve learned knowledge but also reveals shortcomings in addressing higher-level, real-world questions and fine-grained inferences. These findings indicate the need for further fine-tuning to address the capabilities of LLM in bioinformatics tasks. To our knowledge, BioPars is the first application of LLM in Persian medical QA, especially for generating long answers. Evaluation of four selected medical QA datasets shows that BioPars has achieved remarkable results compared to comparative approaches. The model on BioParsQA achieved a ROUGE-L score of 29.99, which is an improvement over GPT-4 1.0. The model achieved a BERTScore of 90.87 with the MMR method. The MoverScore and BLEURT values were also higher in this model than the other three models. In addition, the reported scores for the model are MoverScore=60.43 and BLEURT=50.78. It should be noted that BioPars is an ongoing project, and all resources related to continued development will be made available and can be accessed via the following GitHub repository: \url{https://github.com/amirap80/BioPars}.

\textbf{Keywords:} Biomedical Language Models - Bioinformatics Evaluation - Data-Driven Knowledge Extraction - Benchmarking LLMs -Life Sciences Applications

\end{abstract}
\begin{document}

\flushbottom
\maketitle
%
%
\thispagestyle{empty}

\section{Introduction}

Early benchmarks in NLP were usually oriented towards specific, relatively simple tasks. As Large Language Models (LLMs) outgrew the constraints of previous, small-scale architectures, researchers extended their testing scope to encompass progressively sophisticated abilities. Large-scale benchmarks such as MMLU \cite{hendrycks2020measuring}, BIG-bench \cite{bigb}, and C-Eval \cite{huang2023c} seek to integrate many different NLP challenges in order to provide an overall evaluation of LLMs. Concurrently, various scientific fields have developed their own discipline-specific tests—varying from geoscience \cite{deng2024k2} to radiology \cite{demner2016preparing} and ocean science \cite{bi2023oceangpt}—to meet certain needs of each field and prove the broader context of application for LLMs.
Some recent evaluation frameworks point towards new skills that become more prominent with model size—e.g., reasoning, advanced math \cite{hendrycks2020measuring}, and code generation \cite{chen2021evaluating}. While benchmarks now challenge how LLMs comprehend and use text information in a broad scope of fields, such as medicine and biology \cite{singhal2022large} \cite{bakhshandeh2023benchmarking}, they might provide only bounded comprehension of expert activities. This gap in assessment constrains our capacity to compare the domain-specific strengths of the models, and reliance on multiple-choice formats can be missing the broader set of sophisticated capabilities these systems possess. In this paper, we concentrate on examining the higher-level proficiencies of foundational models within bioinformatics.

Bioinformatics has been one especially intriguing area for applying Large Language Models (LLMs) to biological research \cite{zhang2023applications} \cite{buzatu2023applications} \cite{piccolo2023evaluating}. There has been a variety of research exploring different facets in this intersection, from computer science applications—e.g., medical text processing \cite{li2023comparative}, public health data analysis \cite{yang2023paperdemic}, GNN benchmarking \cite{jha2022prediction}—to biological and medical inquiries, e.g., KVPLM \cite{zeng2022deep}, DNA-Pretrain \cite{luo2023improving}, proteome benchmarks \cite{le2023leveraging} \cite{rao2020transformer} \cite{hallee2023protein}, and peptide property prediction \cite{guntuboina2023peptidebert}. Despite these attempts, much of the LLM community has concentrated on medical text processing, with comparatively limited channels for generative AI to meaningfully assist specialized bioinformatics research. Nonetheless, the convergence of bioinformatics and LLMs promises significant advancements in scientific discovery and technological innovation.

Bioinformatics is a field that combines biology and computational science to gather, examine, and interpret high-level biological data, with a special focus on genomics and molecular biology \cite{bayat2002science}. The emergence of large language models (LLMs) has significantly transformed how scientists manage biological data, opening up new avenues for discovery. Accordingly, mounting demand is for a full-coverage benchmark that such models could be evaluated based on how able they are to solve central bioinformatics tasks. Such a standard would enhance the capacity of current LLMs to facilitate sequence analysis, clinical phenotype prediction, and the discovery of biological insights more effectively.
In order to fill the knowledge gap between the development of LLM and biomedical text mining-specific requirements of the Persian language, we introduce BioPars—an open but powerful testing platform for thoroughly evaluating state-of-the-art reasoning and domain knowledge. Using this platform, we benchmark popular state-of-the-art LLMs, like ChatGPT, Llama, and Galactica. While these models are excellent at multiple-choice type situations, they perform poorly when it comes to more advanced, professional environments. When asked to respond to very domain-specific questions, their responses turn out to be not of tangible utility. Having seen these results, it's clear that targeted training on authentic Persian biomedical data is essential in order to set LLMs properly in viable applications, rather than solely utilizing zero-shot or generic text-generating capabilities.
Eventually, in addition to serving as a standard benchmark, we see BioPars as a multifunctional, modular platform. For practical research as well as application, this approach allows for a deeper, multidimensional evaluation of large language models' comprehension and utilization of domain-specific knowledge. It also helps developers by demystifying each model's areas of strength in the bioinformatics domain, which will ultimately pave the way to more efficient support to researchers in the field.

\section{Related Work}
AI methodologies may benefit all patient-care stages, from initial screening to recovery post-surgery \cite{Farhadi2025Transforming}. Recent biomedical research increasingly blends data-centric analytics with interactive, simulation-based methods. An illustrative case is the MATLAB/Simulink experiential module on vestibular-system dynamics developed by Farhadi Nia et al., which lets learners tune physiological parameters and observe real-time balance responses. Neuromorphic-event cameras coupled with a lightweight deep neural network have likewise enabled real-time positional tracking of moving objects in augmented-reality workspaces while sharply reducing latency and power consumption \cite{Norcéide2024AR}. Such examples highlight the diverse methodological landscape that BioPars seeks to complement with its Persian-language biomedical LLM layer \cite{Nia2023segment}. In a different data-driven domain, logarithmic-polynomial modelling of quality-rate convex hulls has demonstrated high accuracy, with low RMSE and high $R^{2}$ values; these findings suggest that the convex hull of one codec can predict the performance of others and thereby streamline adaptive-streaming workflows \cite{Farhadi2024Explore}.

The growing volume of biomedical corpora and domain-specific AI models has revolutionized medical NLP and text mining. However, biomedical text processing is still sparsely researched in Persian, underscoring the importance of establishing pre-trained Persian biomedical language models, with BioPars as an example. Large public datasets such as the eICU Collaborative Research Database (eICU-CRD) have propelled critical-care NLP by providing high-resolution structured clinical data that enable predictive analytics and decision support~\cite{pollard2018eicu}. Likewise, the MedDialog corpus—3.4 M Chinese and 0.26 M English dialogues—has been used to train conversational medical AI, demonstrating how multiturn dialogue corpora boost language-model performance~\cite{zeng2020meddialog}. Yet most pre-trained models are built on English or Chinese data, leading to poor domain adaptation for Persian. Moreover, the Multi-Scale Attentive Interaction Networks (MSAIN) framework, which improves Chinese medical QA via biGRUs, CNNs, and attention layers, suggests that language-specific model architectures are fundamental~\cite{zhang2018multi}. Therefore, dedicated Persian biomedical corpora and models must be developed.

A similar lesson emerges from medical computer vision: specialized architectures consistently outperform off-the-shelf models. Evaluation on a diverse dataset of breast-ultrasound (BUS) and mammography images shows that the U-Net model surpasses a pretrained Segment-Anything (SAM) architecture in accurately identifying and segmenting both benign and malignant tumours~\cite{Ahmadi2023segment}. U-Net remains superior even for irregular shapes, weak boundaries, and high heterogeneity. These findings reinforce the broader principle that domain-tailored solutions—whether for images or text—yield significant performance gains, motivating the design of BioPars as a Persian-specific biomedical LLM.

To resolve this disparity, both the evaluation benchmark and fine-tuning of biomedical LLMs demand a critical multilingual and domain-specific approach. On one hand, MultiMedQA provides a robust framework for assessing LLM clinical reasoning, utilizing six medical QA datasets in conjunction with HealthSearchQA to enable structured evaluations of biomedical text comprehension and AI safety  \cite{singhal2023large}. On the other hand, human evaluations indicate that even efficient models such as Flan-PaLM generate potentially harmful responses in medical question-answering, underscoring the urgent need for risk-mitigation strategies and culturally sensitive training corpora. Additionally, the Indiana Chest X-ray Collection has proven valuable for biomedical retrieval tasks by aligning radiology images with annotated text reports \cite{demner2016preparing}. With continued exploration in Persian biomedical NLP, datasets such as eICU-CRD, MedDialog, and MultiMedQA will play a pivotal role in developing specialized Persian medical corpora, driving advancements in biomedical text mining, clinical entity recognition, and domain-specific machine translation \cite{zeng2020meddialog} \cite{zhang2018multi}\cite{singhal2023large}\cite{demner2016preparing} . These facts reinforce the critical demand for Persian-specific biomedical datasets and fine-tuned language models. Thus, BioPars is positioned to mitigate medical text mining challenges in low-resource languages.

BioMedGPT \cite{luo2023biomedgpt} introduces a multimodal generative pre-trained transformer designed to bridge the language of life (e.g., molecules, proteins) with human natural language. The model, BioMedGPT-10B, integrates molecular graphs (encoded via GraphMVP) and protein sequences (encoded via ESM-2) with natural language by aligning their feature spaces using a fine-tuned biomedical language model (BioMedGPT-LM-7B), derived from Llama2-Chat-7B and trained on 4.2 million biomedical articles from S2ORC. Evaluated on biomedical QA (MedMCQA, PubMedQA, USMLE), molecule QA (ChEBI-20), and protein QA (UniProtQA) tasks, BioMedGPT-10B outperforms general-purpose models like ChatGPT and Llama2, achieving human-expert-level accuracy on PubMedQA and significant improvements in cross-modal tasks through curated datasets (PubChemQA, UniProtQA). The results highlight its potential in accelerating drug discovery and biomedical research. BioinspiredLLM \cite{luu2024bioinspiredllm} presents a conversational large language model (LLM) fine-tuned on 1,034 peer-reviewed articles in biological and bio-inspired materials mechanics. Built on the Orca-2-13b architecture, the model employs Q-A processed datasets to enhance dialogue coherence and integrates retrieval-augmented generation (RAG) to access external sources and trace references. Evaluated on a specialized 100-question exam, BioinspiredLLM achieved 82\% accuracy (rising to 90\% with chain-of-thought prompting and higher with RAG), outperforming base models in knowledge recall (e.g., mechanical trends in horse hoof hydration) and hypothesis generation (e.g., predicting jackfruit thorn mechanics later validated experimentally). The model also demonstrated utility in collaborative workflows with generative AI tools (e.g., Stable Diffusion) to design bio-inspired materials, highlighting its role in accelerating interdisciplinary research by bridging domain knowledge and enabling creative, data-driven insights.

BioMistral \cite{labrak2024biomistral} emerges as an open-source collection of medical language models built on Mistral 7B Instruct, refined through extensive training on PubMed Central’s biomedical research (3 billion tokens). Tested across 10 specialized medical question-answering benchmarks—including clinical knowledge, genetics, and anatomy—the models surpassed leading open-source alternatives like MedAlpaca and PMC-LLaMA, achieving 50.3\% accuracy in few-shot evaluations and 57.3\% after task-specific tuning, rivaling proprietary tools such as GPT-3.5. Innovations like parameter merging (SLERP/DARE) and memory-efficient quantization boosted adaptability, while multilingual testing (7 languages) revealed strengths and limitations tied to translation quality. While excelling in English medical reasoning, gaps in calibration and cross-lingual consistency highlight avenues for future refinement, positioning BioMistral as a versatile yet evolving resource for healthcare NLP applications.

\cite{guo2021automated} tackles the challenge of efficiently extracting detailed chemical reaction information from scientific journals, aiming to replace the slow, labor-intensive manual methods currently in use. The researchers developed a two-step system: first, identifying key reaction products using specialized neural networks (achieving 76\% precision), and then extracting supporting elements such as catalysts and solvents with 79\% accuracy. By training their models on a small set of annotated examples and leveraging chemistry-specific pretraining, the team was able to navigate the unstructured nature of journal text and capture intricate experimental details. This approach offers a practical solution for processing complex, non-standardized descriptions, making it a valuable tool for speeding up drug development and improving chemical database curation as the volume of scientific literature continues to grow. MatSciBERT \cite{gupta2022matscibert} is a language model specialized for materials science, trained on a corpus of 285 million words from over 150,000 journal articles spanning fields such as ceramics, alloys, and cement. It was developed to overcome the challenges that generic models like SciBERT face with domain-specific terminology. Built on a Transformer architecture, MatSciBERT starts with SciBERT’s weights and then undergoes additional pretraining focused on materials science content. In evaluations, it achieved a Macro-F1 score of 65.92\% on the fine-grained NER task (using the SOFC-Slot dataset), which is about 4\% better than SciBERT, and set new performance standards. It also excelled in classifying glass-related abstracts with a 93.57\% F1 score and showed strong results in relation extraction. The pretrained weights and code are now publicly available to support further developments in materials-specific NLP applications.

\cite{shetty2023general} presents a comprehensive pipeline for automatically extracting material property data from polymer literature, addressing the challenge of mining valuable information from the rapidly growing corpus of materials science publications. The authors curated a dataset of 2.4 million materials science abstracts—of which around 650,000 were polymer-relevant—and manually annotated 750 abstracts using a custom ontology covering eight entity types. They then fine-tuned a BERT-based model, MaterialsBERT (initialized from PubMedBERT), for named entity recognition and relation extraction tasks, achieving superior performance compared to other baseline models on three out of five NER datasets. Using this pipeline, they extracted approximately 300,000 material property records from 130,000 abstracts in just 60 hours with a single GPU, demonstrating its efficacy in accelerating data curation and enabling the discovery of key trends in applications like polymer solar cells, fuel cells, and supercapacitors.

\cite{zhao2024chemdfm} introduces ChemDFM, a large language model (LLM) specialized for chemistry, built by fine-tuning LLaMa-13B through domain pre-training (34B tokens from 3.8M papers and 1.4K textbooks) and instruction tuning (2.7M task-specific prompts). Evaluated on tasks like molecule recognition, property prediction, reaction analysis, and text-based molecule design, ChemDFM outperforms major open-source LLMs (e.g., LLaMa-2, Galactica) and even surpasses GPT-4 in many benchmarks despite its smaller size. It demonstrates robust performance in free-form dialogues, assisting with chemical research tasks such as paper comprehension and experimental design, validated through both quantitative metrics (e.g., 73\% accuracy in SMILES-to-formula tasks) and qualitative real-world scenario testing. The model's weights and code are open-sourced to foster further research.

\cite{zhang2024chemllm} introduces ChemLLM, the first large language model (LLM) specifically designed for chemistry, alongside ChemData—a dataset of 7 million instruction-tuning QA pairs derived from structured chemical databases (e.g., PubChem, ChEMBL)—and ChemBench, a benchmark with 4,100 multiple-choice questions spanning nine core chemical tasks (e.g., molecular property prediction, retrosynthesis). Built on InternLM2-Base-7B, ChemLLM undergoes two-stage tuning: general capability enhancement followed by domain-specific training on ChemData. Results show ChemLLM matches GPT-4’s performance on chemical tasks, achieves state-of-the-art results on ChemBench, and outperforms similar-sized models (e.g., LLaMA2, Mistral) in general benchmarks (MMLU, C-Eval) and specialized NLP tasks like chemical literature translation, demonstrating robust domain expertise while retaining broad linguistic proficiency. 
\cite{yu2024llasmol} presents LlaSMol, a suite of large language models tailored for chemistry using the SMolInstruct dataset—a carefully curated collection that covers 14 chemistry tasks, including name conversion, molecule generation, and retrosynthesis, with more than 3 million samples. The team fine-tuned several base models (Mistral, Llama 2, Code Llama, and Galactica) with LoRA, resulting in models that not only outperform GPT-4 and Claude 3 Opus on key chemistry tasks but also achieve state-of-the-art results on benchmarks like molecule captioning and property prediction. Among these, LlaSMol-Mistral stood out as the top performer, underscoring how domain-specific tuning using a diverse and well-curated dataset like SMolInstruct can produce models that are both highly effective for specialized tasks and versatile enough for broader applications.

\cite{chen2024pharmgpt} introduces PharmaGPT, a suite of domain-specific large language models (13B and 70B parameters) tailored for bio-pharmaceutical and chemical domains, trained on a diverse corpus of specialized data (patents, research papers, clinical reports, exams like NAPLEX, and multilingual resources). The models, built on LLaMA with extended vocabulary and continued pretraining, underwent instruction fine-tuning and reinforcement learning with human feedback (RLHF) to enhance domain expertise. Evaluations on benchmarks such as the North American Pharmacist Licensure Examination (NAPLEX), Chinese Pharmacist Exams, and biomedical translation tasks demonstrated that PharmaGPT outperforms general-purpose models (e.g., GPT-3.5, GPT-4) in accuracy and precision, achieving scores of 66-76\% on NAPLEX and surpassing GPT-4 in medical translation tasks, despite having fewer parameters. Results highlight its efficacy in specialized pharmaceutical and chemical applications.
\cite{pollard2018eicu} introduces the eICU Collaborative Research Database, a freely accessible, multi-center database containing high-granularity clinical data from over 200,000 ICU admissions across 208 US hospitals (2014–2015), including vital signs, lab results, medications, diagnoses, treatments, and APACHE IV severity scores. Designed to support critical care research, the database facilitates tasks such as developing machine learning algorithms, clinical decision tools, and benchmarking studies. Structured into 26 relational tables (e.g., patient, lab, medication), the dataset is rigorously deidentified and validated for integrity, with data completeness varying by hospital (e.g., vital signs available for >94\% of patients). Hosted on PhysioNet, access requires ethical training and a data-use agreement, enabling researchers to explore ICU outcomes, treatment patterns, and predictive modeling while ensuring patient privacy.

In conjunction with the development of LLMs, there has been a growing demand for domain-specific evaluation benchmarks, such as those in biomedical text mining contexts. There are a number of general-purpose benchmarks, such as SciEval and Xiezhi, which assess scientific knowledge and interdisciplinary reasoning \cite{sun2024scieval} \cite{gu2024xiezhi}; Consequently, they don't address the challenges in biomedical NLP, where domain-specific terminology, abbreviations, and context-sensitive interpretation are crucial. Moreover, as indicated by C-EVAL, Chinese-language models require linguistic adaptation. Thus, it illustrates that non-English contexts are challenging even for advanced models in terms of higher-order reasoning \cite{huang2024c}. This is of particular relevance to the mining of biomedical texts in Persian, since there is a serious lack of both annotated biomedical datasets and pre-trained language models in Persian. Furthermore, AGIEval, evaluating reasoning by human-standardized exams, stresses that though LLMs do great in structured question-answering tasks, they underperform in fields which require deep, domain-specific knowledge and thus reinforce the need for specialized biomedical benchmarks \cite{zhong2023agieval}.

As a benchmark that focuses on domain expertise, BIOINFO-BENCH offers a rich evaluation of LLM performance in bioinformatics-related fields including sequence analysis, disease classification, and analysis of biomedical data \cite{chen2023bioinfo}. The findings from BIOINFO-BENCH shed light on the fact that even efficient models like ChatGPT and Galactica struggle with real-world biomedical problem-solving, suggesting that general pretraining on extensive corpora is inadequate for specialized biomedical applications. In addition, SCIENCEQA provides a multimodal evaluation framework that underscores the importance of interpretable AI in biomedical NLP by utilizing not only text but also diagrams and domain-specific scientific knowledge \cite{lu2022learn}. These benchmarks collectively reveal that biomedical language models call for domain-specific pretraining, fine-tuning on comprehensive medical corpora, and evaluation metrics tailored to sophisticated biomedical reasoning rather than general natural language understanding. Accordingly, the limitations identified by MMLU in tasks requiring mathematical and scientific problem-solving indicate that LLMs still struggle with structured reasoning and are therefore not optimal for biomedical NLP applications that demand clinical decision support and automated diagnosis \cite{hendrycks2020measuring}.
To mitigate these challenges, precise benchmarking systems that employ reasoning-intensive biomedical questions and domain-specific adaptation are essential for biomedical LLMs. The AI2 Reasoning Challenge (ARC) has demonstrated that LLMs perform poorly on inference-based science questions, suggesting that domain knowledge falls short without robust reasoning capabilities \cite{clark2018think}. This insight is particularly relevant to Persian biomedical NLP, where the scarcity of large-scale annotated datasets further complicates model development and evaluation. Therefore, future research should focus on creating biomedical benchmarks for Persian language models, incorporating clinical, pharmaceutical, and genetic data while maintaining interpretability and robustness in text mining applications. BioPars strives to bridge this gap by taking cues from both general and biomedical benchmarks to develop a Persian biomedical LLM that enhances biomedical text mining, supports clinical decision-making, and advances biomedical research in resource-limited contexts.






\section{BIOPARS-BENCH}

This section explains the method and goals of a detailed medical database collected from Persian medical books, journals, and teaching materials. The collection includes many medical subjects, from basic to advanced healthcare fields, and is a useful tool for researchers, doctors, and tech developers creating medical tools in Persian.
The dataset is compiled from authoritative Persian medical literature, including textbooks, clinical guidelines, and peer-reviewed articles.

\begin{Summary}{Key topics covered include }{firstsummary}

\begin{itemize}
    \item \textbf{Hematology: }Iron Deficiency Anemia, Hemophilia, Hematology, Blood Donation
\item	\textbf{Neurology \& Genetics:} Parkinson’s Disease Genetics, Migraine
\item	\textbf{Infectious Diseases: }COVID-19 Prevention Guidelines, Influenza
\item	\textbf{Pharmacology:} Antibiotics, Diuretic Drugs, Anti-Nausea Medications, Pharmacological Principles
\item	\textbf{Diagnostics \& Imaging: }Radiology, Interpretation of Laboratory Tests
\item	\textbf{Maternal \& Pediatric Health:} Breastfeeding and Infant Nutrition, Neonatal Care
\item	\textbf{Chronic Conditions:} Hypertension Self-Care Education, Autoimmune Disorders
\item	\textbf{Surgical \& Clinical Care:} Burn Management, Hemorrhoid Treatment, Vaginal Infections
\item	\textbf{Nursing \& General Medicine: }Nursing Practices, Comparison of Intravenous (IV) Solutions
\end{itemize}
The dataset includes 20+ core categories with granular subtopics, ensuring broad coverage of clinical, preventive, and therapeutic medical knowledge.
\end{Summary}

\begin{figure*}
\centering
\subfloat[2025 ]{\includegraphics[width = 0.3\textwidth]{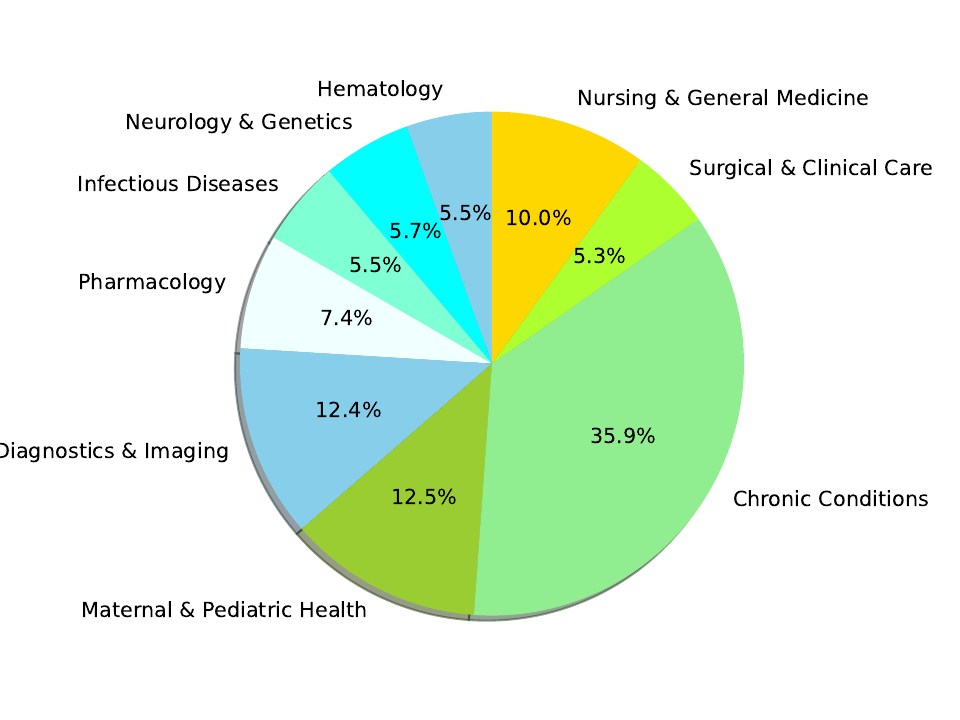}}
\subfloat[2024]{\includegraphics[ width = 0.3\textwidth]{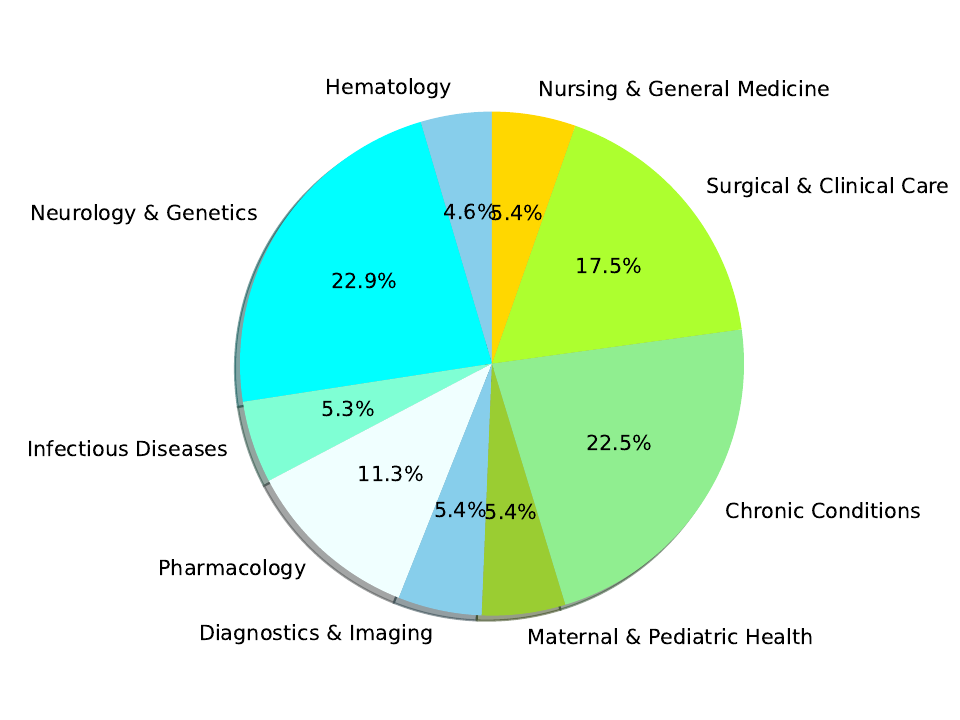}}
\subfloat[2023]{\includegraphics[width = 0.3\textwidth]{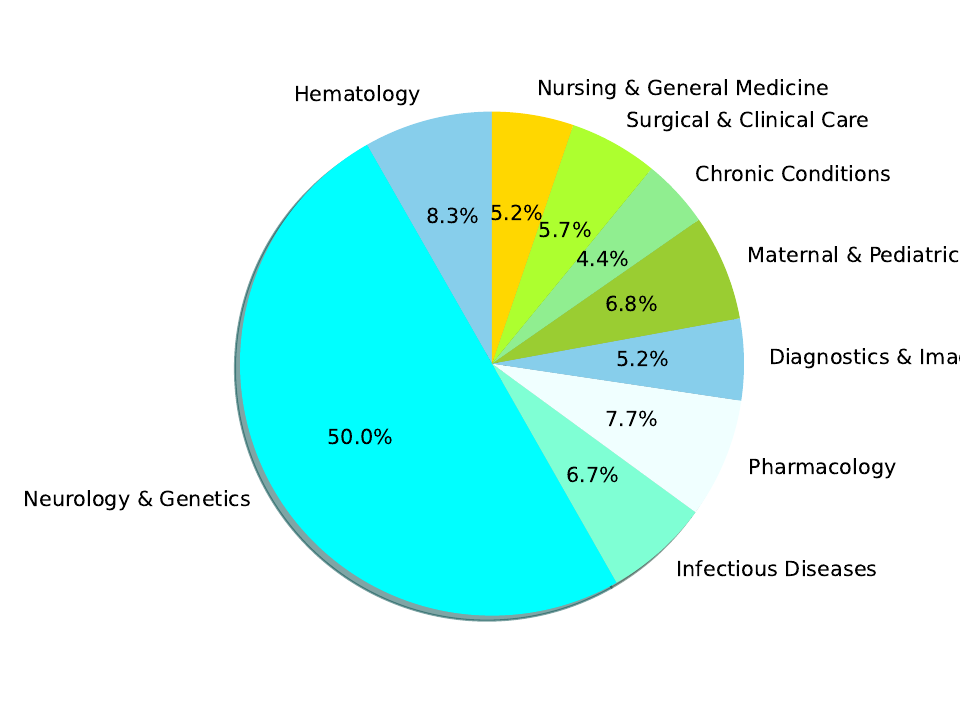}}\\
\subfloat[2022]{\includegraphics[width = 0.3\textwidth]{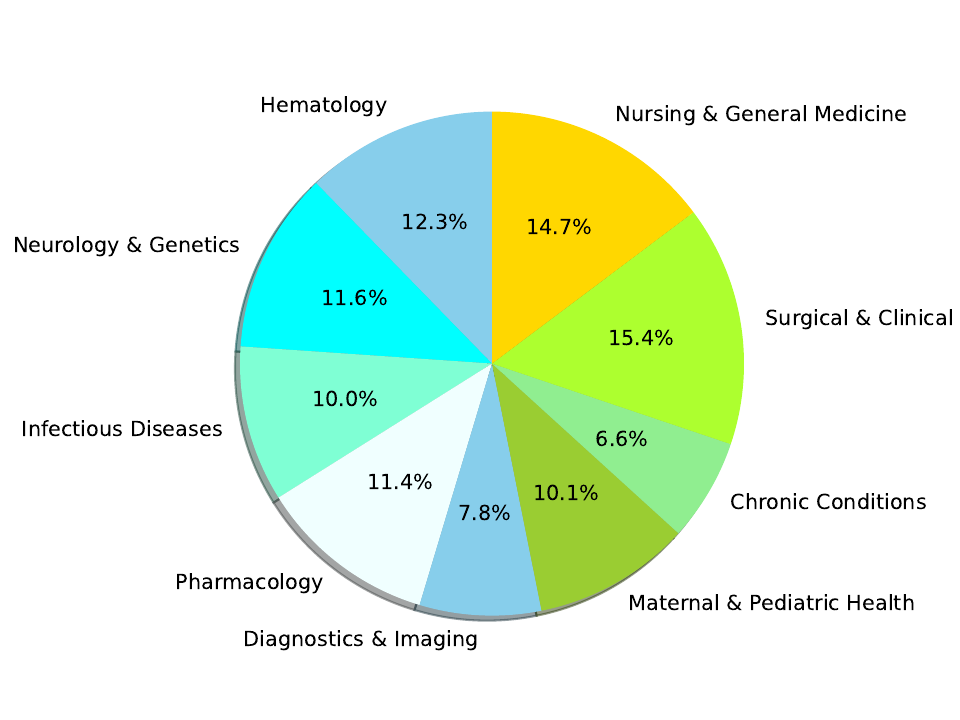}}
\subfloat[2021]{\includegraphics[width = 0.3\textwidth]{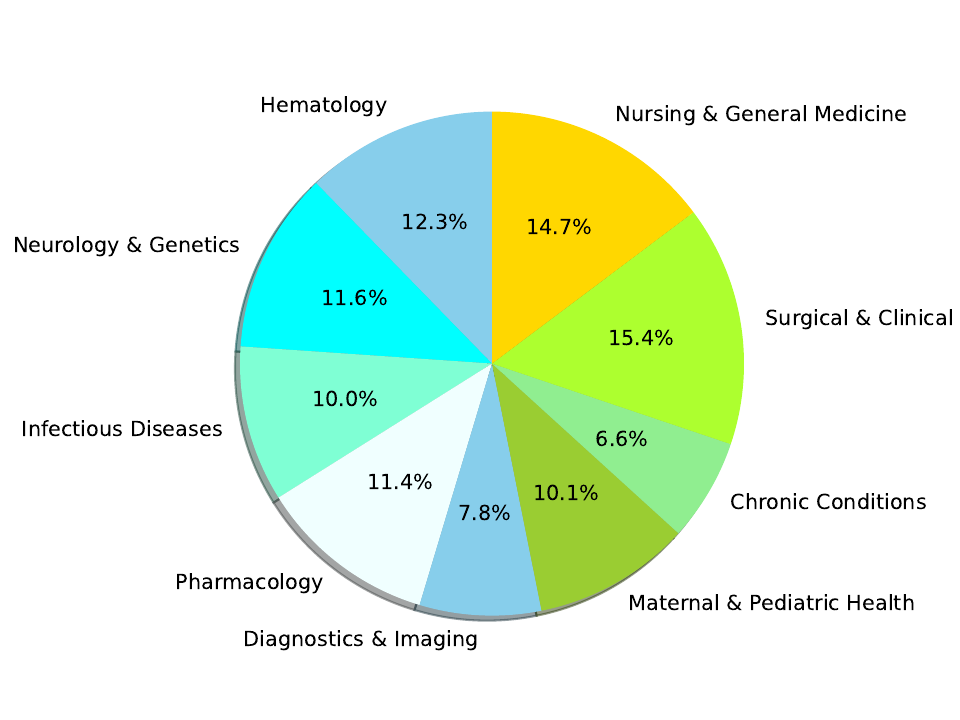}}
\subfloat[2020]{\includegraphics[width = 0.3\textwidth]{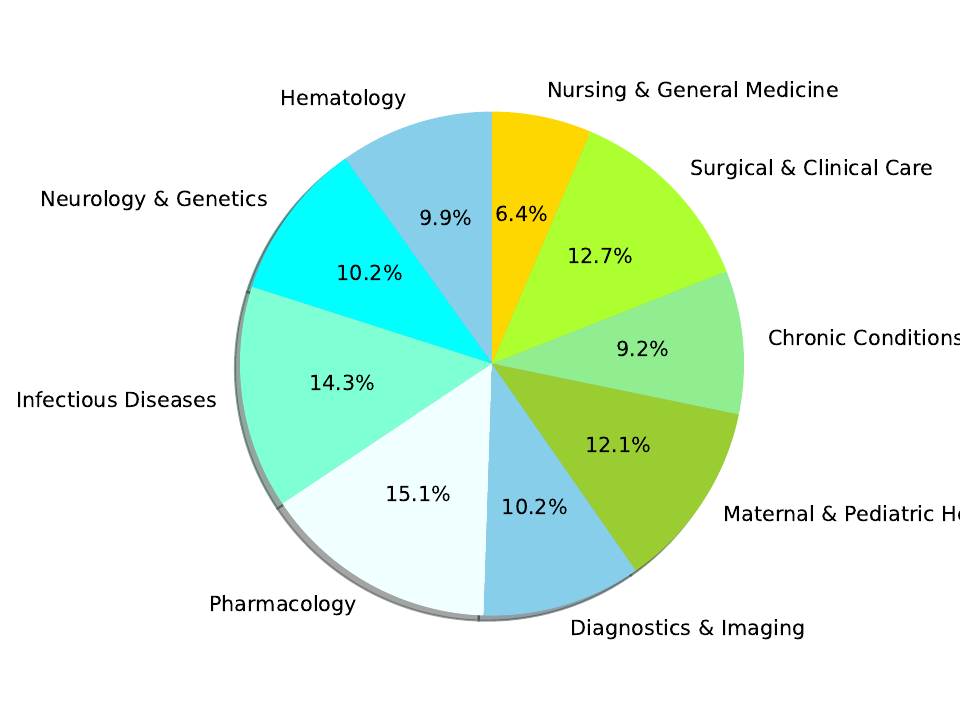}}\\
\subfloat[2019]{\includegraphics[width = 0.3\textwidth]{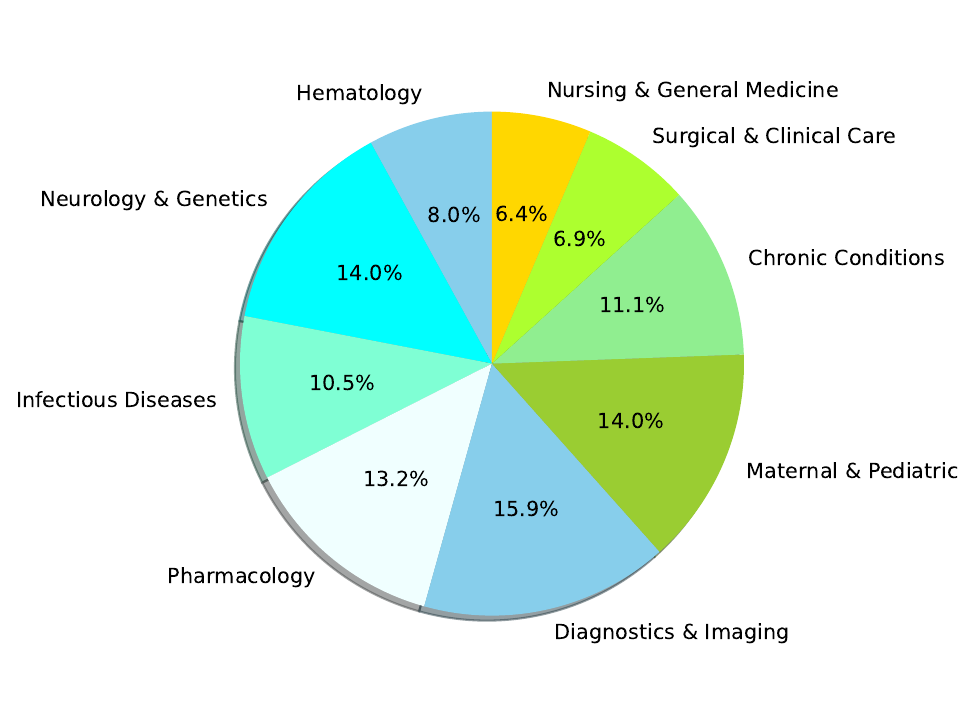}}
\caption{Statistical analysis of the research data set, }
\label{BarChartSahh}
\end{figure*}

Scientific articles in various journals were used to collect text data.
Initially, articles were collected in the Journal of Medical Sciences of the University of Tehran\footnote{\url{https://tumj.tums.ac.ir/}} in 73 intervals between 2018 and 2025.
 Details of the collected articles are given in Table \ref{Table1}. These articles included specialized areas that were mostly registered as innovations and findings in the field of medicine. For more comprehensiveness, the data was collected from 6600 general articles collected from paziresh24\footnote{\url{https://www.paziresh24.com/}}, seebmagazine\footnote{\url{https://seebmagazine.com/}}, Ibn Sina Clinical Medicine Journal\footnote{\url{https://sjh.umsha.ac.ir/}}, and pezeshkyemrooz\footnote{\url{https://www.pezeshkyemrooz.com/}}. Details of general keywords and the number of words collected from scientific articles are given in Table \ref{Table1}.

\begin{table}[]
\caption{The number of words collected from
 scientific articles}
\centering
\begin{tabular}{|l|c|c|c|c|}
\hline

Source                                    & year & \# Interval & \# Article & \# words \\ \hline
\multirow{7}{*}{https://tumj.tums.ac.ir/} & 2025 & 1           & 70         & 67000    \\ \cline{2-5} 
                                          & 2024 & 12          & 127        & 17012    \\ \cline{2-5} 
                                          & 2023 & 12          & 127        & 182243   \\ \cline{2-5} 
                                          & 2022 & 12          & 121        & 19022    \\ \cline{2-5} 
                                          & 2021 & 12          & 108        & 167881   \\ \cline{2-5} 
                                          & 2020 & 12          & 124        & 165900   \\ \cline{2-5} 
                                          & 2019 & 12          & 121        & 171420   \\ \hline
https://www.paziresh24.com/               & -    & -           & 2100       & 8 M      \\ \hline
https://seebmagazine.com/                 & -    & -           & 2000       & 8.7 M    \\ \hline
https://sjh.umsha.ac.ir/                  & -    & -           & 2000       & 10.1 M   \\ \hline
https://www.pezeshkyemrooz.com/           & -    & -           & 500        & 2 M      \\ \hline
\end{tabular}
\label{Table1}
\end{table}

\section{BioParsQA }
\begin{figure*}[!hptb]
\begin{center}
\includegraphics[width=1\textwidth]{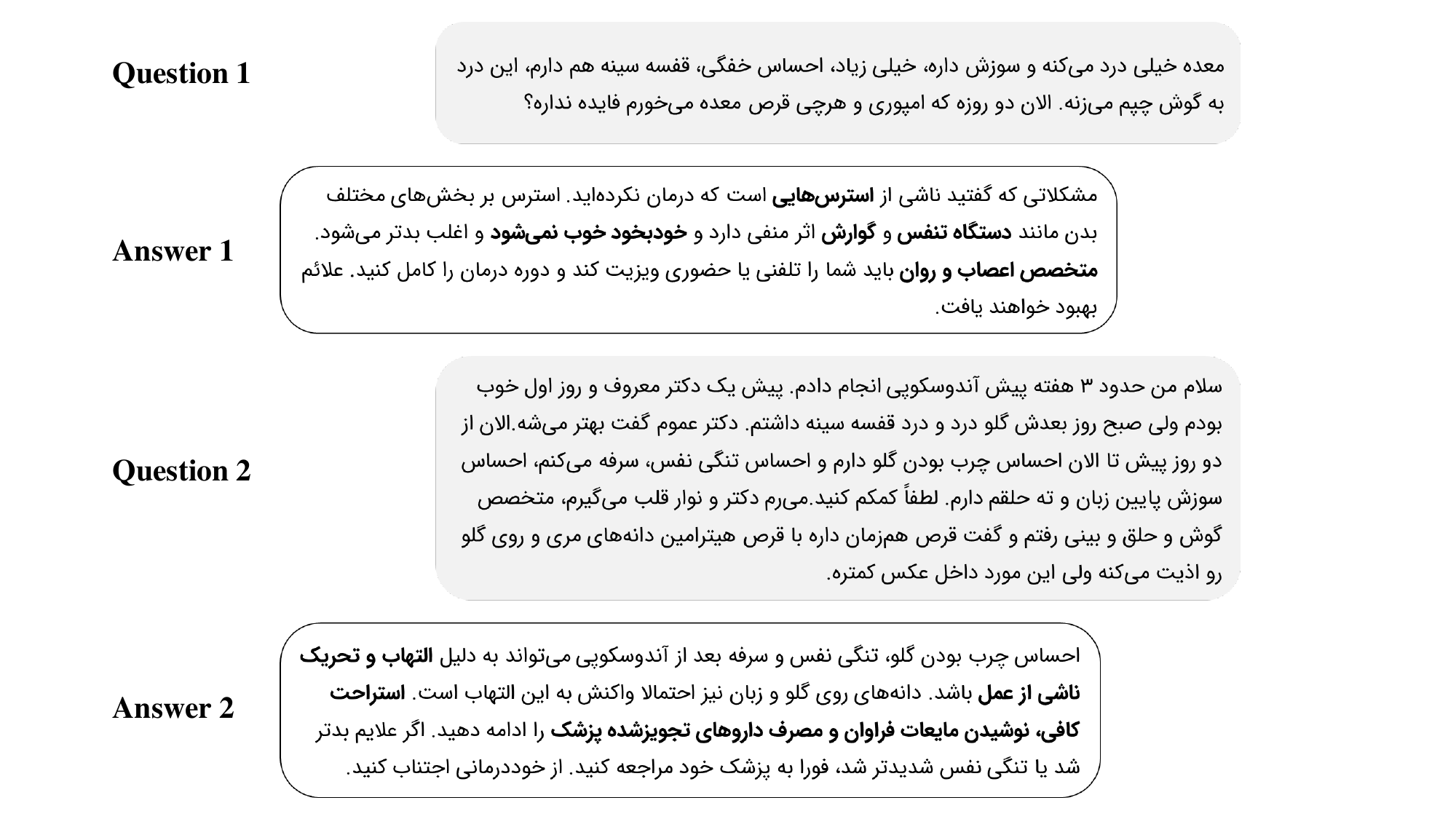}
\caption{A sample of questions and answers provided in BioparsQA.}
\label{FigQA1}
\end{center}
\end{figure*}

To evaluate the proposed LLM, we need a dataset related to the context in which it is used. For this purpose, a set of medical questions and answers was collected. Answering online questions usually allows for errors due to the possibility of answers from non-specialists, which can affect the final result and the final decision. On the other hand, providing a wrong answer has irreparable consequences. For example, the drug Mifepristone is usually provided to treat high blood sugar. This drug causes fetal loss in pregnant women. For this purpose, an attempt was made to use questions and answers that at least two specialist doctors approved. 5231 medical questions and answers were collected. This dataset is publicly available for research purposes. An example of these questions and answers is given in Figure \ref{FigQA1} .

\section{BioPars}

 \subsection{BioPars Encoder: Gated Attention Guided by Moving Average}
Assume that a biomedical token sequence is $X={x_1, x_2,...,x_n}\in R^{n*d}$. Following a single encoder block, the representations are $Y={y_1,...,y_n}\in R^{n*d}$.
Ordinary self-attention is enhanced by BioPars using a damped exponential moving average (EMA), which introduces a significant temporal bias. The outcome is then modulated by lightweight gating.

\subsection{High-Dimensional Damped EMA}
An expansion matrix $\beta \in R^{d*h}$ first lifts each scalar component of a token vector to an h-dimensional latent space:

\begin{equation}
    u_t^j=\beta_jx_{t,j}
\end{equation}
Next, an element-wise damped EMA is applied:
\begin{equation}
    h_t^j=\alpha_j \bigodot u_t^j+(1-\alpha_j \bigodot \delta_j)\bigodot h_{t-1}^j
\end{equation}
Using $\eta \in R^{d*h}$, the hidden state is projected back to the original width:
\begin{equation}
    y_{t,j}=\eta_j^Th_t^j
\end{equation}

\subsection{The Gated Attention Layer}
Short-range dependencies are already encoded in the shared context vector provided by the EMA output. This vector is used by BioPars to construct reset- and update-gates, which are then combined with scaled dot-product attention and merged via a residual path to produce a small but expressive block.
\subsubsection{Restrictions}
\begin{itemize}
  \item \textbf{Local context}: The cost is decreased from O(n 2) to O(nc) by dividing queries, keys, and values into chunks of length c; however, inter-chunk interactions are weakened.
  \item \textbf{Drift in configuration}: A fragmented design space resulted from earlier versions' requirement for distinct normalization options and gating strategies for various data types.
  \item \textbf{Unknown Scalability:} Stability in multi-million token and billion-parameter contexts has not yet been tested.
\end{itemize}
\subsection{BioPars with Large-Scale, Long-Context Improvements}
BioPars implements a number of architectural improvements to get around those problems:
\subsubsection{Complex-Domain EMA, or C-EMA}
EMA can be extended to the complex plane by considering it a diagonal state-space model. Each actual decay is substituted with a complex coefficient by BioPars:

\begin{equation}
    h_t^j=\alpha_j(cos \theta_j+ i sin \theta_j)\bigodot u_t^j+(1-\alpha_j\bigodot \delta_j)(cos \theta_j+i sin \theta_j)\bigodot h_{t-1}^j
\end{equation}

\begin{equation}
    y_{t,j}=Re(\eta_j^Th_t^j)
\end{equation}
$\alpha,\delta \in R^{d*h}$  are the real number parameters same as in EMA.
Different from EMA, $\eta \in \mathbb{C}^{d \times h}$ in CEMA are complex numbers. $\theta_j \in \mathbb{R}^h$, $j \in \{1,2,\ldots,d\}$ are the $h$ arguments.

\subsubsection{Normalization of Timesteps}
While drift along the sequence axis is ignored, LayerNorm equalizes variance across features. Thus, for small feature groups, BioPars calculates a cumulative mean and variance, producing TimestepNorm, which stabilizes activations without disclosing future tokens—essential for autoregressive pre-training.
\subsubsection{Normalized Attention}
An $l2$-normalized shared vector $\hat{Z}$ is the source of queries and keys:
\begin{equation}\label{eq:cema_output}
  X' = \mathrm{CEMA}(X), \quad X' \in \mathbb{R}^{n \times d}
\end{equation}

\begin{equation}\label{eq:z_norm}
  Z = X'W_z + b_{z'}, \quad Z' = \frac{Z}{\|Z\|}, \quad Z' \in \mathbb{R}^{n \times z}
\end{equation}

\begin{equation}\label{eq:q}
  Q = k_q \odot Z' + \mu_q, \quad Q \in \mathbb{R}^{n \times z}
\end{equation}

\begin{equation}\label{eq:k}
  K = k_k \odot Z' + \mu_k, \quad K \in \mathbb{R}^{n \times z}
\end{equation}

Where $Q$ and $K$ are computed using the normalized shared representation $Z'$ instead of $Z$.

Then, the attention operation is defined as:

\begin{equation}\label{eq:attention}
  O = \mathrm{softmax}(QK^T)V, \quad O \in \mathbb{R}^{n \times v}
\end{equation}

After which a softmax is sufficient—no explicit scale or SiLU is required. The outcome is less susceptible to saturation and more stable.

\subsubsection{Pre-Norm Two-Hop Residual}
BioPars returns the feed-forward branch straight to the block input rather than via a parameter-heavy gated residual:
\begin{equation}\label{s}
  \hat{Y}=Attention(Norm(X))+X
\end{equation}
\begin{equation}\label{s}
  Y=FFN(Norm(\hat{Y}))+\hat{Y}=FFN(Norm(\hat{Y}))+Attention(Norm(X))+X
\end{equation}
Removing extra gates while maintaining variance control even at ~7 B parameters.

\subsubsection{Parallelism }
In addition to data, tensor, and pipeline parallelism, chunk-wise attention opens up a sequence-parallel dimension. Pre-training on Persian biomedical corpora with contexts well beyond typical Transformer limits is made possible by the fact that only the final C-EMA state and TimestepNorm statistics traverse devices, and that these transfers fully overlap with computation.

\section{Evaluation Metrics }
Four criteria were used to evaluate the results: BERTScore\cite{zhang2019bertscore}, ROUGE\cite{lin2004rouge}, Mover Score\cite{zhao2019moverscore}, and BLEURT\cite{sellam2020bleurt}. The following is a general definition of these criteria.
\subsection{ROUGE}
ROUGE is a set of automatic tools to check summary quality by comparing computer-made summaries with human-written ones. It measures overlap in four main ways\cite{lin2004rouge}:

\begin{itemize}
    \item \textbf{ROUGE-N}: Counts matching word groups (like single words or word pairs)
    
    \item \textbf{ROUGE-L}: Finds longest word sequences in order (even with gaps)
    
    \item \textbf{ROUGE-W}: Rewards consecutive matches more than scattered ones
    
    \item \textbf{ROUGE-S}: Checks word pairs with gaps (e.g., "dog cat" matches "dog ran after cat")
\end{itemize}

All versions balance\cite{lin2004rouge}:
\begin{itemize}
    \item {Coverage}: How much human content is included
    \item {Relevance}: How much computer content is useful
\end{itemize}

ROUGE focuses more on coverage (using setting $\beta$) and is widely used because it matches human judgments well, while being faster and cheaper than manual checks.
\begin{itemize}
    \item \textbf{ROUGE-N}:$
\text{ROUGE-N} = \frac{\sum_{S \in \{\text{Refs}\}} \sum_{\text{gram}_n \in S} \text{Count}_{\text{match}}(\text{gram}_n)}{\sum_{S \in \{\text{Refs}\}} \sum_{\text{gram}_n \in S} \text{Count}(\text{gram}_n)}
$
\item \textbf{ROUGE-L}
\begin{align*}
R_{\text{LCS}} &= \frac{\text{LCS}(X,Y)}{m}, \quad 
P_{\text{LCS}} = \frac{\text{LCS}(X,Y)}{n} \\
\text{ROUGE-L} &= \frac{(1+\beta^2)R_{\text{LCS}}P_{\text{LCS}}}{R_{\text{LCS}} + \beta^2 P_{\text{LCS}}}
\end{align*}
\item \textbf{ROUGE-W}
\begin{align*}
R_{\text{WLCS}} &= f^{-1}\left(\frac{\text{WLCS}(X,Y)}{f(m)}\right) \\
P_{\text{WLCS}} &= f^{-1}\left(\frac{\text{WLCS}(X,Y)}{f(n)}\right) \\
\text{ROUGE-W} &= \frac{(1+\beta^2)R_{\text{WLCS}}P_{\text{WLCS}}}{R_{\text{WLCS}} + \beta^2 P_{\text{WLCS}}}
\end{align*}
\item \textbf{ROUGE-S}
\begin{align*}
R_{\text{SKIP2}} &= \frac{\text{SKIP2}(X,Y)}{\binom{m}{2}}, \quad 
P_{\text{SKIP2}} = \frac{\text{SKIP2}(X,Y)}{\binom{n}{2}} \\
\text{ROUGE-S} &= \frac{(1+\beta^2)R_{\text{SKIP2}}P_{\text{SKIP2}}}{R_{\text{SKIP2}} + \beta^2 P_{\text{SKIP2}}}
\end{align*}
\item \textbf{ROUGE-SU:} $\text{ROUGE-SU} = \text{ROUGE-S} + \text{Unigram-F1}$
\end{itemize}

\subsection{BERTScore}

BERTScore helps evaluate how good computer-generated text is by comparing it to human-written versions. Instead of just counting matching words, it uses special language processing from systems like BERT to understand meaning connections. The tool gives three main ratings\cite{zhang2019bertscore}:

\begin{itemize}
    \item \textbf{Recall (R\_BERT)}: How much important info from human text is kept
    \item \textbf{Precision (P\_BERT)}: How much computer text is useful
    \item \textbf{F1 Score (F\_BERT)}: Balanced combination of both scores
    \item Uses middle layers of BERT (layer 9) for better meaning matching
    \item Adjusts scores to 0-1 scale using random text comparisons
\end{itemize}

\subsubsection*{Formulas}
\begin{itemize}
    \item \textbf{Basic Scores}
\begin{equation*}
R_{\text{BERT}} = \frac{1}{|x|} \sum_{x_i \in x} \max_{\hat{x}_j \in \hat{x}} x_i^\top \hat{x}_j
\end{equation*}

\begin{equation*}
P_{\text{BERT}} = \frac{1}{|\hat{x}|} \sum_{\hat{x}_j \in \hat{x}} \max_{x_i \in x} x_i^\top \hat{x}_j
\end{equation*}

\begin{equation*}
F_{\text{BERT}} = \frac{2 \cdot P_{\text{BERT}} \cdot R_{\text{BERT}}}{P_{\text{BERT}} + R_{\text{BERT}}}
\end{equation*}
 \item \textbf{IDF Weighting (Optional)}: $
\text{idf}(w) = -\log\left(\frac{1}{M}\sum_{i=1}^M \mathbb{I}[w \in x^{(i)}]\right)$
 \item \textbf{IDF-Weighted Recall} $
R_{\text{BERT}} = \frac{\sum_{x_i \in x} \text{idf}(x_i) \cdot \max_{\hat{x}_j \in \hat{x}} x_i^\top \hat{x}_j}{\sum_{x_i \in x} \text{idf}(x_i)}
$
\end{itemize}

\subsection{MoverScore Metric}
MoverScore evaluates text generation quality by computing the semantic distance between system outputs ($\bm{x}$) and reference texts ($\bm{y}$) using \textbf{contextualized embeddings} and \textbf{Earth Mover's Distance (EMD)}~\cite{zhao2019moverscore}. Unlike n-gram metrics, it:

\begin{itemize}
\item Encodes text using contextual models (BERT/ELMo)
\item Aggregates layer-wise representations via power means or routing
\item Uses IDF-weighted $n$-gram embeddings to emphasize content words
\item Solves optimal transport to find minimal alignment cost
\end{itemize}

\subsubsection*{Formulation}
For $n$-gram sequences $\bm{x}^n = (x_1^n,\dots,x_m^n)$ and $\bm{y}^n = (y_1^n,\dots,y_k^n)$:

\begin{equation}
\text{MoverScore}(\bm{x}^n, \bm{y}^n) := \min_{\bm{F} \in \mathbb{R}_+^{m \times k}} \underbrace{\sum_{i=1}^m \sum_{j=1}^k \bm{C}_{ij} \bm{F}_{ij}}_{\text{Total Transportation Cost}} 
\end{equation}

Subject to:
\begin{align}
\bm{F}\mathbf{1} &= \bm{f}_{\bm{x}^n} \quad \text{(Source marginal)} \\
\bm{F}^\top\mathbf{1} &= \bm{f}_{\bm{y}^n} \quad \text{(Target marginal)}
\end{align}

Where $\bm{C}_{ij} = \|E(\bm{x}_i^n) - E(\bm{y}_j^n)\|_2$ is the Euclidean distance between $n$-gram embeddings, $E(\bm{x}_i^n) = \sum_{t=i}^{i+n-1} \text{idf}(x_t) \cdot \phi(\bm{z}_{t,1:L})$ is the IDF-weighted aggregated embedding, $\phi(\cdot)$ is the layer aggregation (e.g., power means $\frac{1}{L}\sum_{l=1}^L \bm{z}_l^{p}$), and  $\bm{f}_{\bm{x}^n}[i] = \frac{\sum_{t=i}^{i+n-1}\text{idf}(x_t)}{Z}$ is the normalized $n$-gram weights.

\subsubsection*{Key Variations}
\begin{equation}
\text{WMD}_n = \begin{cases} 
\text{Word-level } (n=1) \\ 
\text{Bigram-level } (n=2) \\
\text{Sentence-level } (n=\text{max})
\end{cases}, \quad
\text{SMD} = \|E(\bm{x}) - E(\bm{y})\|_2
\end{equation}

\subsection{BLEURT Metric}
BLEURT is a learned evaluation metric for text generation that leverages BERT embeddings and multi-task pre-training to align with human judgments. It first generates synthetic data by perturbing Wikipedia sentences (via mask-filling, backtranslation, and word dropping) and trains on diverse signals (BLEU, ROUGE, textual entailment, etc.) to capture lexical, syntactic, and semantic deviations. The model is then fine-tuned on human-rated data. BLEURT's pre-training ensures robustness to domain shifts and sparse supervision\cite{sellam2020bleurt}.

\subsubsection*{Formulation}
The pre-training objective combines multiple supervision signals:
\begin{equation}
\ell_{\text{pre-train}} = \frac{1}{M} \sum_{m=1}^M \sum_{k=1}^K \gamma_k \ell_k(\bm{\tau}_k^m, \hat{\bm{\tau}}_k^m)
\end{equation}

Where  $\gamma_k$ is task-specific weighting parameters, $\ell_k$ is loss function (regression $\ell_2$ or cross-entropy), $\bm{\tau}_k^m$ is supervision signal for task $k$ on example $m$, $\hat{\bm{\tau}}_k^m$ is model prediction from BERT's [CLS] embedding, $M$ is Number of synthetic training examples, and $K$ is  the number of pre-training tasks.

Evaluation metrics in response to ranking methods such as Similarity Ranking (Sim), Answer Extension Ranking (AE), and Maximum Marginal Relevance (MMR) Ranking were also used, along with Re-Ranking (RR), which is on top of Similarity Ranking.
\section{Result}

In order to determine a suitable performance standard, BioPars is compared to three
of the more popular large language models that differ in terms of size, licensing, and
intended use case.
\begin{itemize}
    \item \textbf{ChatGPT}\cite{williams2023evaluating}: ChatGPT is a closed-source LLM lineage that consists of commercial models
with proprietary internal architecture and training data that are only available through
APIs. Although they are unable to view or change model weights, end users are able
to generate text. There will occasionally be truncated or redacted biomedical outputs
because responses are processed through proprietary trust-and-safety layers,
illustrating the trade-off between transparency and ease of access.
    \item \textbf{Llama 7B}\cite{touvron2023llama}: This 7 billion-parameter version of Meta AIs Llama 2 suite is a lightweight, fully open-source foundation model. Because of its permissive license,
researchers can modify or expand the model for specialized domains, which makes
it a viable option for laboratory-scale studies. Despite the existence of larger Llama
checkpoints (13 B–70 B), the 7 B edition is featured here because it is
computationally affordable and falls at the threshold where emergent reasoning first
becomes noticeable.
    \item \textbf{Galactica 30B}\cite{taylor2022galactica}: Galactica 30B is a 30-billion-parameter LLM that was pre-trained on scientific text by Metas Papers with Code. It is optimized for academic tasks like
entity extraction, formal summarization, question answering across disciplines, and
structured citation prediction. Galacticas generative fluency occasionally falls short
of more general-purpose models, despite its domain specificity, providing an
intriguing contrast between linguistic versatility and specialization.
\end{itemize}
\begin{table}[]
\centering
\begin{tabular}{|l|c|c|c|c|}
\hline
Dataset      & Sentence (Q) & Word (Q) & Sentence (A) & Word (A) \\ \hline
LiveQA       & 1.15         & 14.76    & 6.96         & 141.02   \\ \hline
ExpertQA-Bio & 1.26         & 21.69    & 6.18         & 184.38   \\ \hline
ExpertQA-Med & 1.37         & 22.19    & 5.96         & 180.55   \\ \hline
MedQA        & 1.02         & 7.36     & 3.38         & 71.48    \\ \hline
BioParsQA    & 1.42         & 25.34    & 5.9          & 176.09   \\ \hline
\end{tabular}
\caption{
The average statistic is the data in each data set, where Q represents Question and A represents Answer (data details are taken from \cite{yang2024kg}.)}
\label{TData1}
\end{table}

We also used four other medical QA datasets to evaluate the proposed method and protocol. The details of these datasets are given in Table \ref{TData1}. These datasets include the following: 
\begin{enumerate}
    \item \textbf{LiveQA\cite{abacha2017overview}:} Contains health questions submitted to the National Library of Medicine. This dataset includes a training set with 634 QA pairs and a test set with 104 QA pairs used for evaluation. 
    \item \textbf{ExpertQA\cite{malaviya2023expertqa}:} A high-quality long-form QA dataset with 2177 questions in 32 fields, with answers validated by experts. Among them, 504 medical questions and 96 biology questions were used for model evaluation. 
    \item \textbf{ MedicationQA\cite{abacha2019bridging}:} Contains 690 consumer-related questions related to drugs with information retrieved from reputable websites and scientific articles.
    
\end{enumerate}

\begin{figure*}
\centering
\subfloat[GPT-4 ]{\includegraphics[width = 0.5\textwidth]{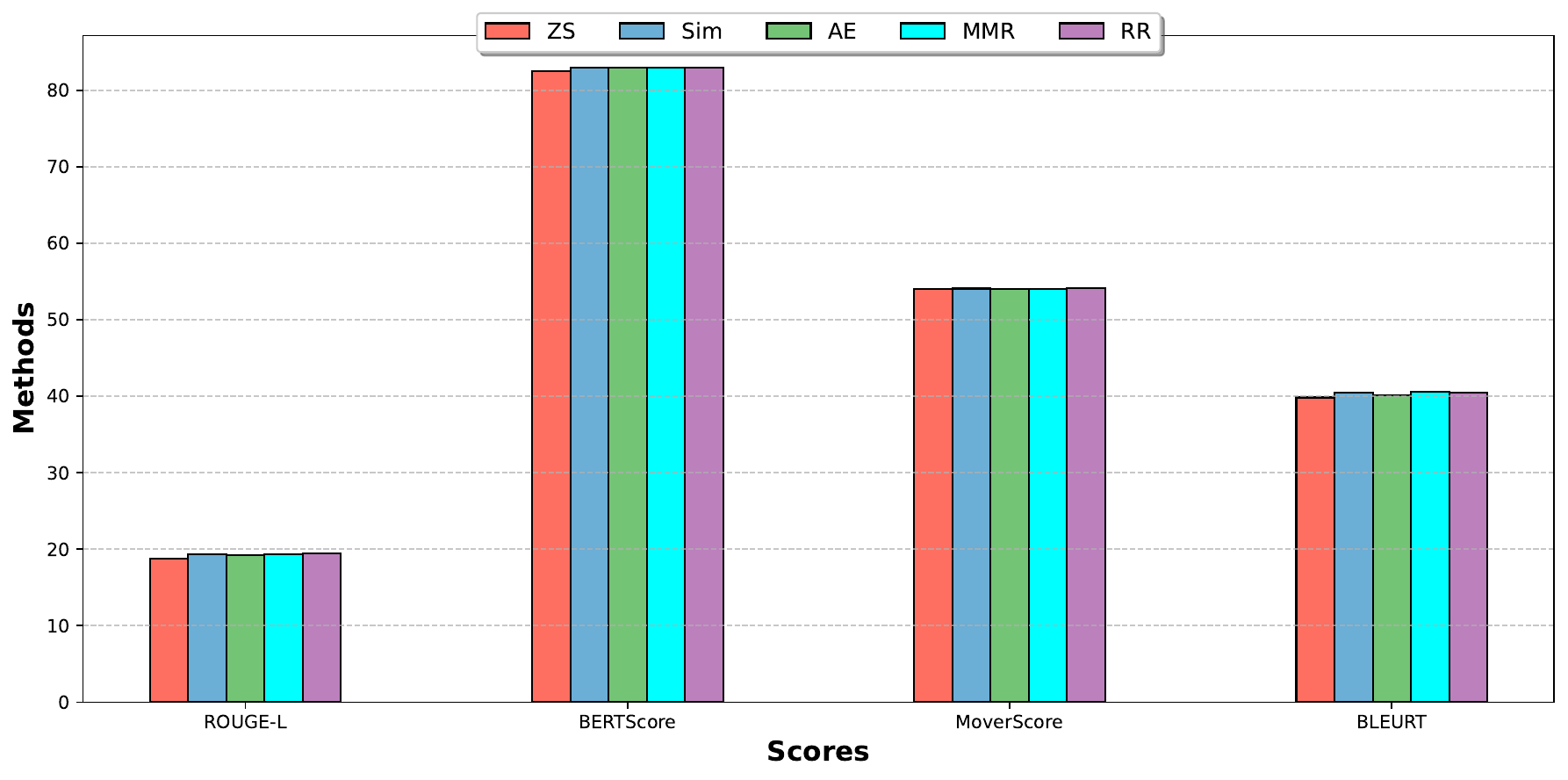}}
\subfloat[LLaMA2-13b]{\includegraphics[ width = 0.5\textwidth]{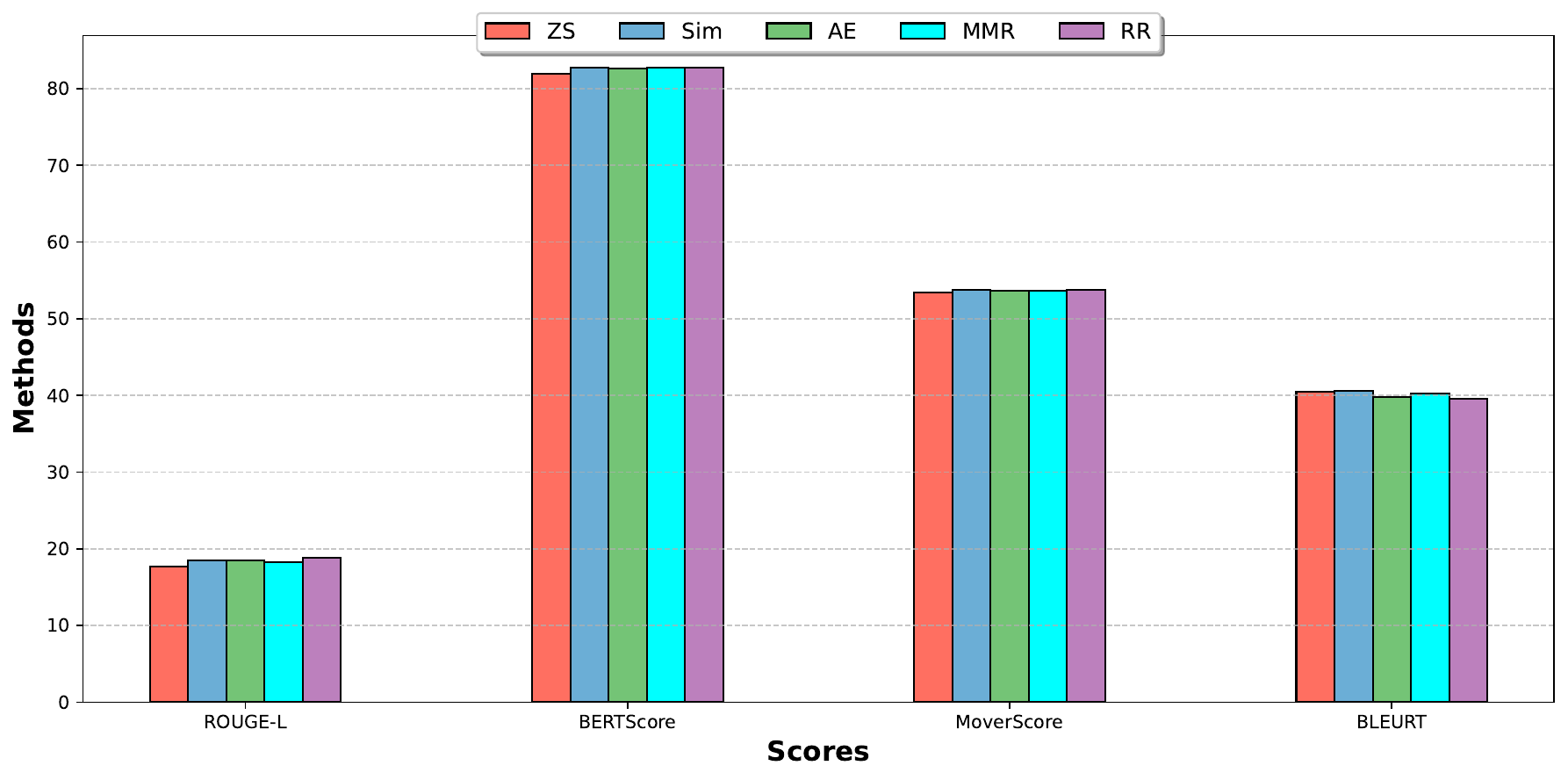}}\\
\subfloat[Galactica 30B]{\includegraphics[width = 0.5\textwidth]{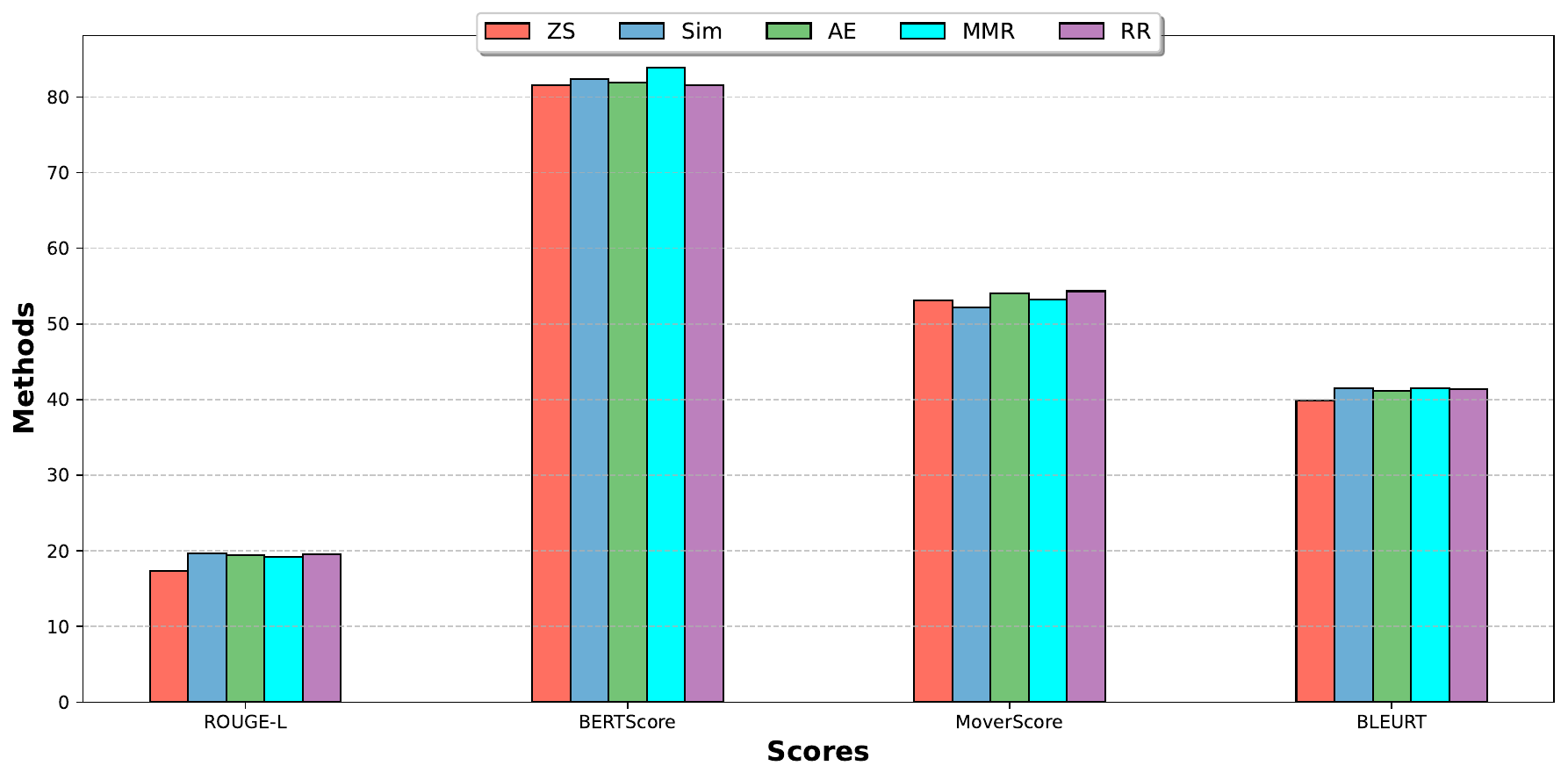}}
\subfloat[BioPars]{\includegraphics[width = 0.5\textwidth]{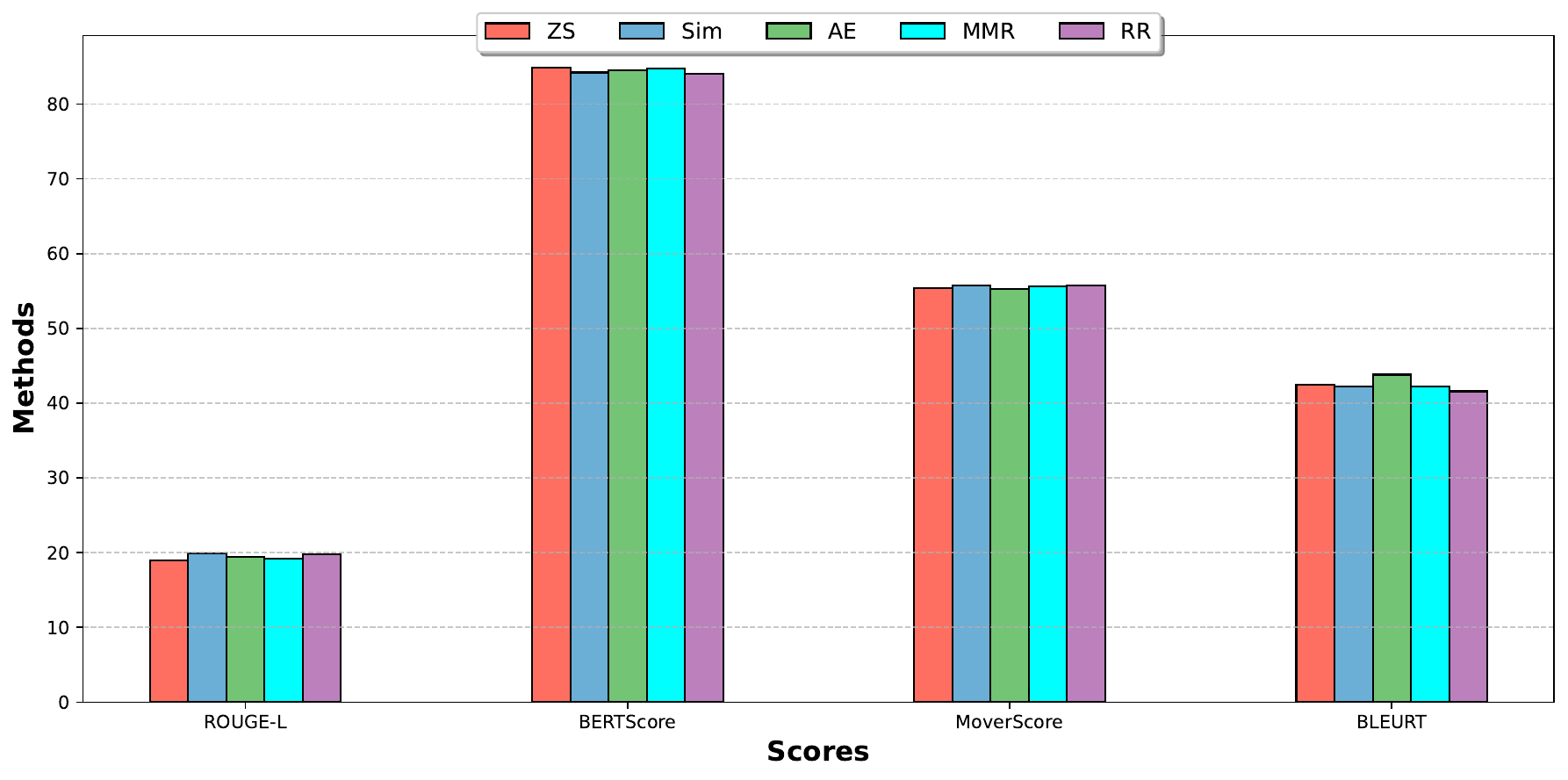}}\\
\caption{Comparative bar plot of different approaches on the LiveQA .}
\label{BarChartLive}
\end{figure*}

In the evaluation of the LiveQA dataset (see Table \ref{TableR1}), the GPT-4 model was able to achieve ROUGE-l=19.44 using the RR method. It also achieved maximum BertScore=83.01, MoverScore=54.11, and BLEURT=40.55. In evaluating the LLaMA2-13b model on the LiveQA dataset, the results were weaker than those of GPT-4. It achieved maximum ROUGE-L=18.83, BertScore=82.79, MoverScore=53.79, and BLEURT=40.59 in the RR, RR, Sim, and Sim evaluation methods. The Galactica 30B model achieved better results in some evaluation criteria than the LLaMA2-13b and GPT-4 models. It achieved maximum ROUGE-L=19.68. It also achieved maximum BERTScore in the three evaluation criteria. MoverScore and BLEURT maximum results were obtained at 83.92, 54.32, and 41.52, respectively. BioPars achieved the best results in the evaluation criteria on the LiveQA data. This model achieved a maximum ROUGE- L=19.8, which performed about 0.40 better than GPT-4. This model also achieved a maximum BERTScore=84.92. The maximum MoverScore and BLEURT values were equal to 55.73 and 43.81, respectively.
The comparative bar plot of different approaches on the LiveQA is shown in Figure \ref{BarChartLive}.

\begin{table}[]
\resizebox{\textwidth}{!}{%
\begin{tabular}{|l|l|llll|llll|llll|llll|}
\hline
Dataset                 & Methods & \multicolumn{4}{c|}{GPT-4}                                                                               & \multicolumn{4}{c|}{LLaMA2-13b}                                                                           & \multicolumn{4}{c|}{Galactica   30B}                                                                      & \multicolumn{4}{c|}{BioPars:}                                                                             \\ \hline
                        &         & \multicolumn{1}{l|}{ROUGE-L} & \multicolumn{1}{l|}{BERTScore} & \multicolumn{1}{l|}{MoverScore} & BLEURT & \multicolumn{1}{l|}{ROUGE- L} & \multicolumn{1}{l|}{BERTScore} & \multicolumn{1}{l|}{MoverScore} & BLEURT & \multicolumn{1}{l|}{ROUGE- L} & \multicolumn{1}{l|}{BERTScore} & \multicolumn{1}{l|}{MoverScore} & BLEURT & \multicolumn{1}{l|}{ROUGE- L} & \multicolumn{1}{l|}{BERTScore} & \multicolumn{1}{l|}{MoverScore} & BLEURT \\ \hline
\multirow{5}{*}{LiveQA} & ZS      & \multicolumn{1}{l|}{18.8}    & \multicolumn{1}{l|}{82.50}     & \multicolumn{1}{l|}{54.02}      & 39.84  & \multicolumn{1}{l|}{17.73}    & \multicolumn{1}{l|}{81.93}     & \multicolumn{1}{l|}{53.37}      & 40.45  & \multicolumn{1}{l|}{17.32}    & \multicolumn{1}{l|}{81.52}     & \multicolumn{1}{l|}{53.11}      & 39.82  & \multicolumn{1}{l|}{18.98}    & \multicolumn{1}{l|}{84.92}     & \multicolumn{1}{l|}{55.35}      & 42.42  \\ \cline{2-18} 
                        & Sim     & \multicolumn{1}{l|}{19.35}   & \multicolumn{1}{l|}{83.01}     & \multicolumn{1}{l|}{54.08}      & 40.47  & \multicolumn{1}{l|}{18.52}    & \multicolumn{1}{l|}{82.78}     & \multicolumn{1}{l|}{53.79}      & 40.59  & \multicolumn{1}{l|}{19.68}    & \multicolumn{1}{l|}{82.32}     & \multicolumn{1}{l|}{52.22}      & 41.45  & \multicolumn{1}{l|}{19.92}    & \multicolumn{1}{l|}{84.23}     & \multicolumn{1}{l|}{55.73}      & 42.21  \\ \cline{2-18} 
                        & AE      & \multicolumn{1}{l|}{19.24}   & \multicolumn{1}{l|}{82.95}     & \multicolumn{1}{l|}{54.04}      & 40.15  & \multicolumn{1}{l|}{18.45}    & \multicolumn{1}{l|}{82.60}     & \multicolumn{1}{l|}{53.70}      & 39.80  & \multicolumn{1}{l|}{19.43}    & \multicolumn{1}{l|}{81.93}     & \multicolumn{1}{l|}{54.04}      & 41.12  & \multicolumn{1}{l|}{19.42}    & \multicolumn{1}{l|}{84.53}     & \multicolumn{1}{l|}{55.23}      & 43.81  \\ \cline{2-18} 
                        & MMR     & \multicolumn{1}{l|}{19.32}   & \multicolumn{1}{l|}{\textbf{82.91}}     & \multicolumn{1}{l|}{54.03}      & 40.55  & \multicolumn{1}{l|}{18.25}    & \multicolumn{1}{l|}{82.70}     & \multicolumn{1}{l|}{53.67}      & 40.22  & \multicolumn{1}{l|}{19.19}    & \multicolumn{1}{l|}{83.92}     & \multicolumn{1}{l|}{53.23}      & 41.52  & \multicolumn{1}{l|}{19.22}    & \multicolumn{1}{l|}{84.72}     & \multicolumn{1}{l|}{55.63}      & 42.24  \\ \cline{2-18} 
                        & RR      & \multicolumn{1}{l|}{\textbf{19.44}}   & \multicolumn{1}{l|}{82.94}     & \multicolumn{1}{l|}{\textbf{54.11}}      & 40.50  & \multicolumn{1}{l|}{18.83}    & \multicolumn{1}{l|}{82.79}     & \multicolumn{1}{l|}{53.72}      & 39.59  & \multicolumn{1}{l|}{19.54}    & \multicolumn{1}{l|}{81.54}     & \multicolumn{1}{l|}{54.32}      & 41.33  & \multicolumn{1}{l|}{19.8}     & \multicolumn{1}{l|}{84.00}     & \multicolumn{1}{l|}{55.71}      & 41.58  \\ \hline
\end{tabular}}
\caption{Automatic evaluation scores on LiveQA : we compare ROUGE-L, BERTScore, MoverScore, BLEURT on different
settings. The superior scores among the same models are highlighted in bold.}
\label{TableR1}
\end{table}

\begin{figure*}
\centering
\subfloat[GPT-4 ]{\includegraphics[width = 0.5\textwidth]{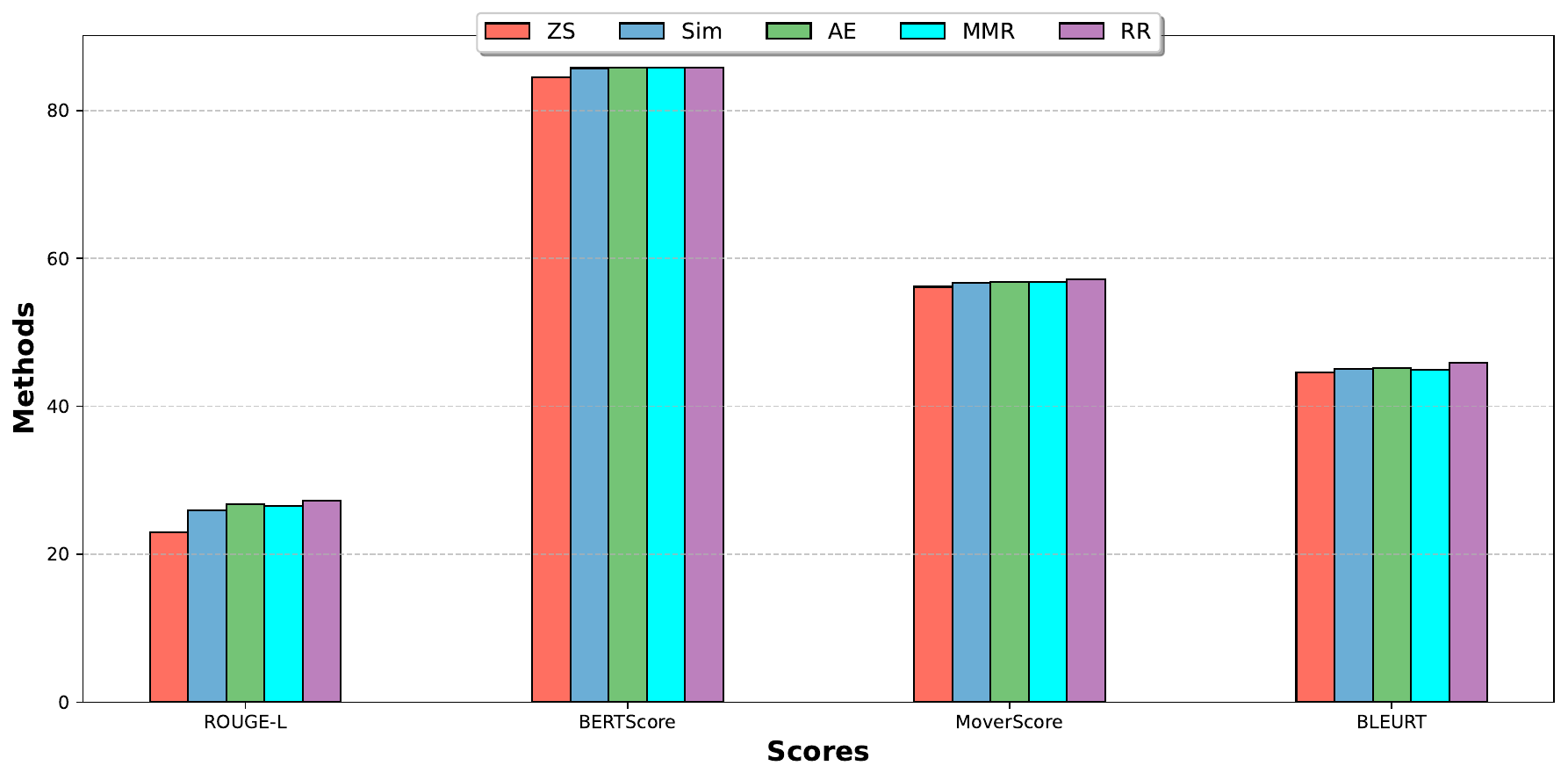}}
\subfloat[LLaMA2-13b]{\includegraphics[ width = 0.5\textwidth]{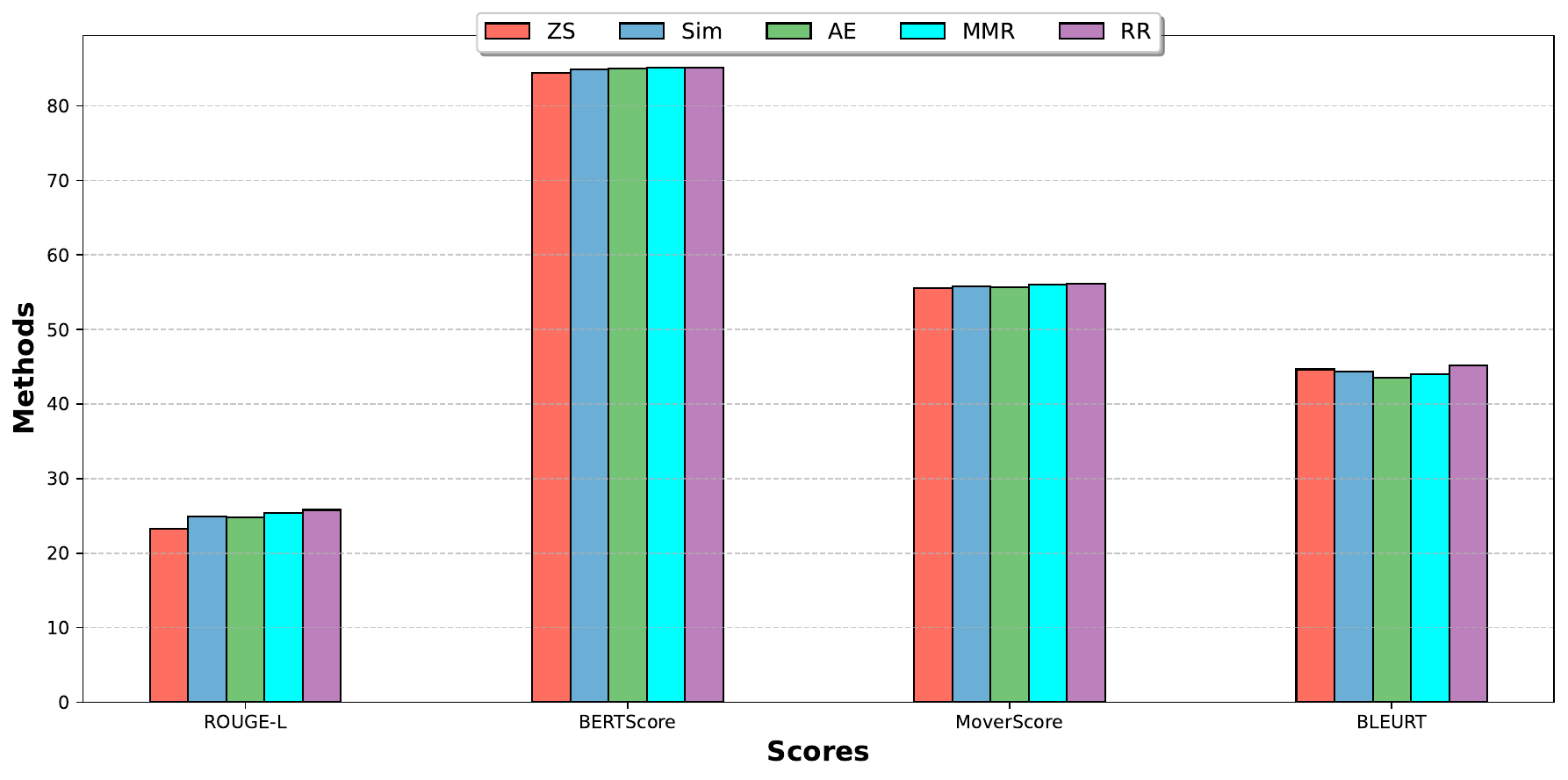}}\\
\subfloat[Galactica   30B]{\includegraphics[width = 0.5\textwidth]{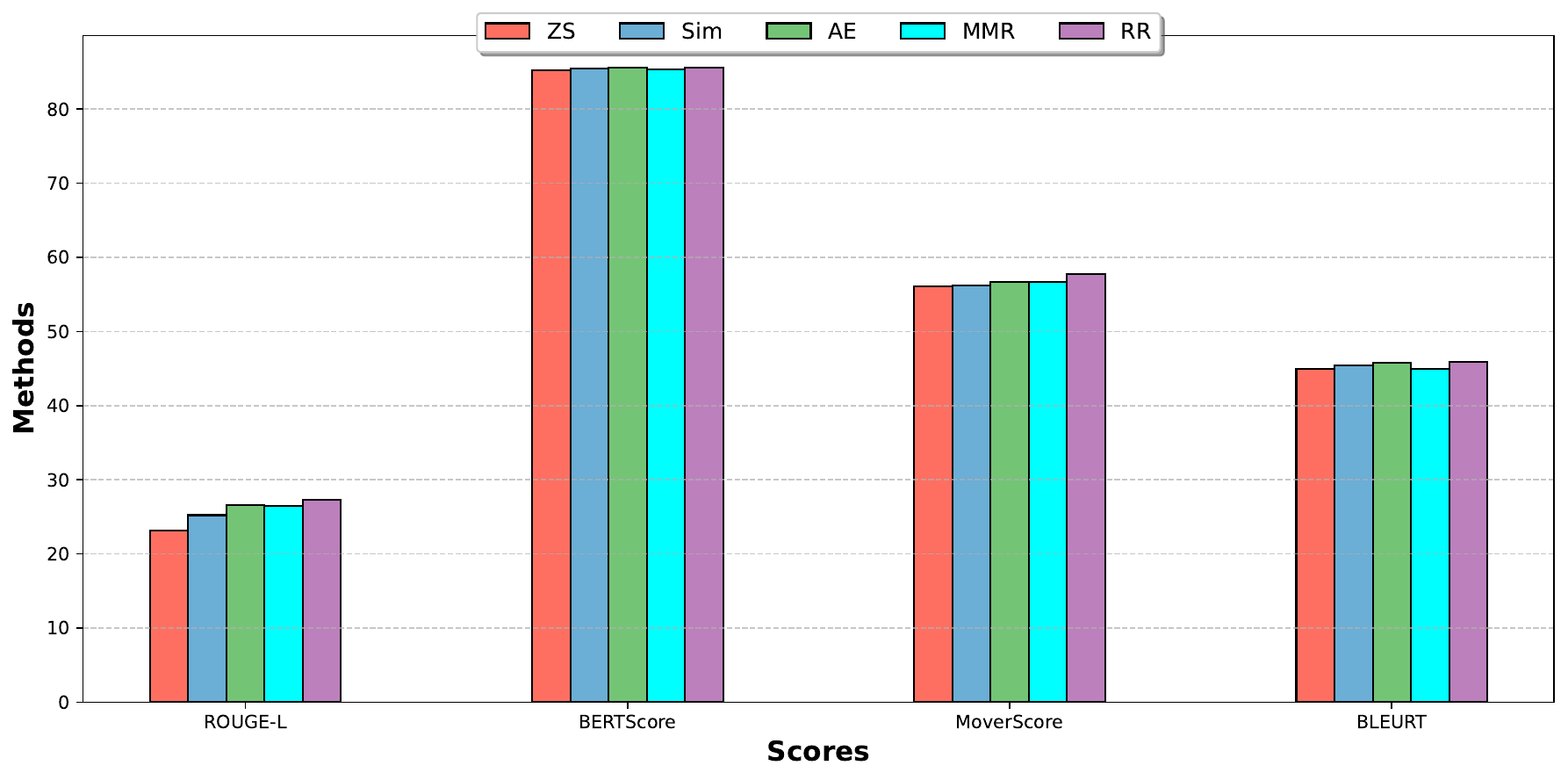}}
\subfloat[BioPars]{\includegraphics[width = 0.5\textwidth]{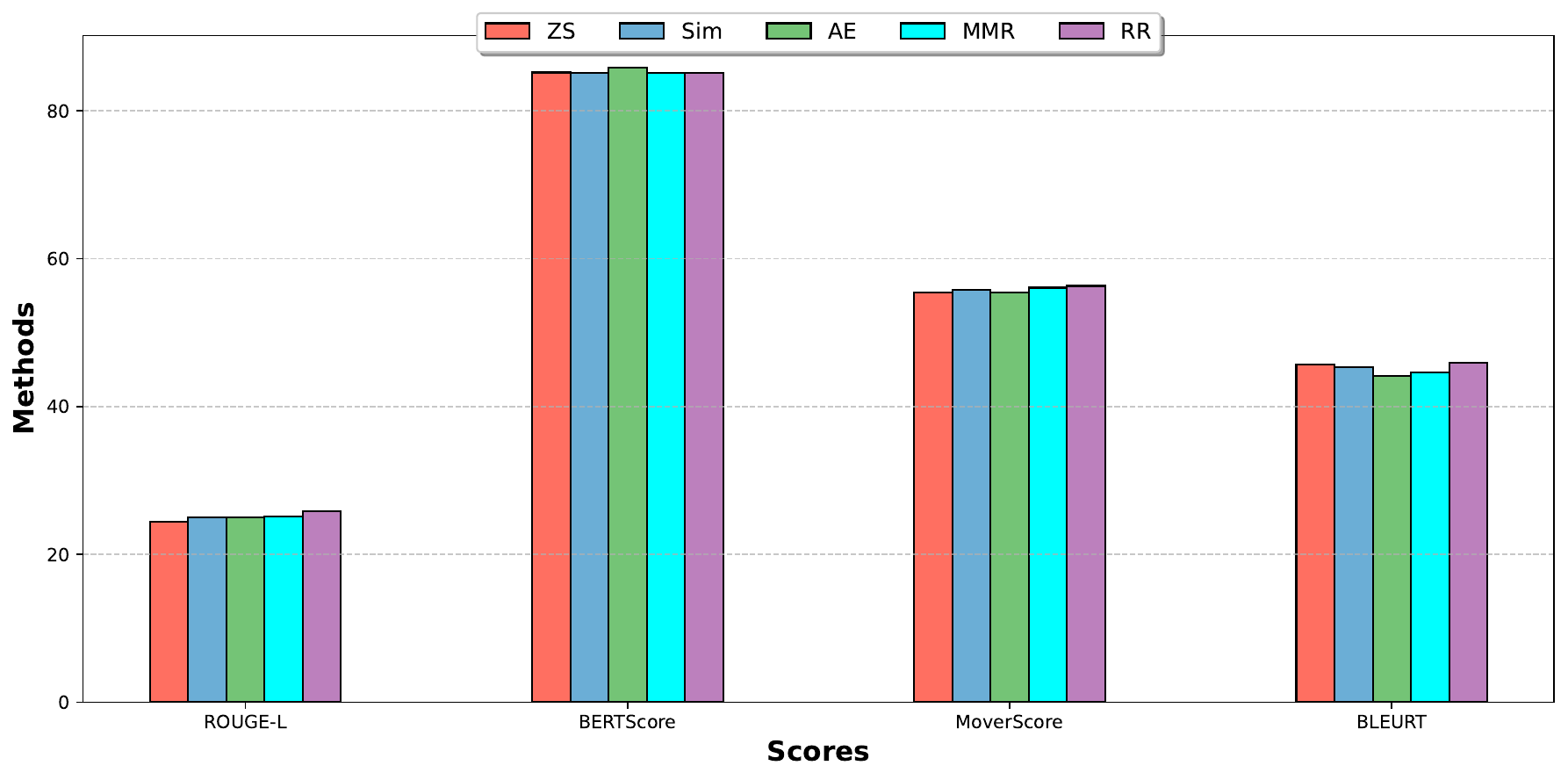}}\\
\caption{comparative bar plot of different approaches on the ExpertQA-Bio.}
\label{BarChartEXBi}
\end{figure*}

The evaluation of the models on the ExpertQA-Bio dataset is also shown in Table \ref{TableR2}. The GPT-4 model achieved ROUGE-L=27.20 on this data with the maximum RR evaluation method. Also, the highest BERTScore, MoverScore, and BLEURT values for this model were 85.83, 56.79, and 45.93, which were obtained with the RR, AE, and MMR methods, respectively. The LLaMA2-13b model presented weaker results than GPT-4. This model achieved ROUGE-L=25.79, BERTScore=85.18, MoverScore=56.17, and BLEURT=45.20 in the RR evaluation method, the highest values reported for this model. Galactica 30B also achieved results close to LLaMA2-13b in some evaluation methods. This model recorded 27.32, 85.65, 56.73, and 45.87 values in the evaluation metrics. BioPars obtained a maximum ROUGE- L = 25.82, which performed worse than the GPT-4 model in this metric. This model also achieved a maximum BERTScore of 85.12, which performed worse than Galactica 30B in other evaluation metrics.
The comparative bar plot of different approaches on the ExpertQA-Bio is shown in Figure \ref{BarChartEXBi}.

\begin{table}[]
\resizebox{\textwidth}{!}{%
\begin{tabular}{|l|l|cccc|cccc|cccc|cccc|}
\hline
Dataset                       & Methods & \multicolumn{4}{c|}{GPT-4}                                                                               & \multicolumn{4}{c|}{LLaMA2-13b}                                                                           & \multicolumn{4}{c|}{Galactica   30B}                                                                      & \multicolumn{4}{c|}{BioPars}                                                                              \\ \hline
                              &         & \multicolumn{1}{c|}{ROUGE-L} & \multicolumn{1}{c|}{BERTScore} & \multicolumn{1}{c|}{MoverScore} & BLEURT & \multicolumn{1}{c|}{ROUGE- L} & \multicolumn{1}{c|}{BERTScore} & \multicolumn{1}{c|}{MoverScore} & BLEURT & \multicolumn{1}{c|}{ROUGE- L} & \multicolumn{1}{c|}{BERTScore} & \multicolumn{1}{c|}{MoverScore} & BLEURT & \multicolumn{1}{c|}{ROUGE- L} & \multicolumn{1}{c|}{BERTScore} & \multicolumn{1}{c|}{MoverScore} & BLEURT \\ \hline
\multirow{5}{*}{ExpertQA-Bio} & ZS      & \multicolumn{1}{c|}{23.00}   & \multicolumn{1}{c|}{84.50}     & \multicolumn{1}{c|}{56.15}      & 44.53  & \multicolumn{1}{c|}{23.26}    & \multicolumn{1}{c|}{84.38}     & \multicolumn{1}{c|}{55.58}      & 44.65  & \multicolumn{1}{c|}{23.11}    & \multicolumn{1}{c|}{85.22}     & \multicolumn{1}{c|}{56.11}      & 45.01  & \multicolumn{1}{c|}{24.44}    & \multicolumn{1}{c|}{85.21}     & \multicolumn{1}{c|}{55.43}      & 45.69  \\ \cline{2-18} 
                              & Sim     & \multicolumn{1}{c|}{25.90}   & \multicolumn{1}{c|}{85.72}     & \multicolumn{1}{c|}{56.73}      & 45.10  & \multicolumn{1}{c|}{24.96}    & \multicolumn{1}{c|}{84.91}     & \multicolumn{1}{c|}{55.83}      & 44.35  & \multicolumn{1}{c|}{25.23}    & \multicolumn{1}{c|}{85.43}     & \multicolumn{1}{c|}{56.22}      & 45.43  & \multicolumn{1}{c|}{25.01}    & \multicolumn{1}{c|}{85.12}     & \multicolumn{1}{c|}{55.84}      & 45.32  \\ \cline{2-18} 
                              & AE      & \multicolumn{1}{c|}{26.78}   & \multicolumn{1}{c|}{85.77}     & \multicolumn{1}{c|}{56.79}      & 45.18  & \multicolumn{1}{c|}{24.84}    & \multicolumn{1}{c|}{84.97}     & \multicolumn{1}{c|}{55.72}      & 43.55  & \multicolumn{1}{c|}{26.54}    & \multicolumn{1}{c|}{85.65}     & \multicolumn{1}{c|}{56.73}      & 45.81  & \multicolumn{1}{c|}{24.96}    & \multicolumn{1}{c|}{85.91}     & \multicolumn{1}{c|}{55.44}      & 44.15  \\ \cline{2-18} 
                              & MMR     & \multicolumn{1}{c|}{26.54}   & \multicolumn{1}{c|}{85.76}     & \multicolumn{1}{c|}{56.77}      & 44.93  & \multicolumn{1}{c|}{25.40}    & \multicolumn{1}{c|}{85.08}     & \multicolumn{1}{c|}{55.98}      & 44.04  & \multicolumn{1}{c|}{26.51}    & \multicolumn{1}{c|}{85.34}     & \multicolumn{1}{c|}{56.65}      & 44.99  & \multicolumn{1}{c|}{25.12}    & \multicolumn{1}{c|}{85.12}     & \multicolumn{1}{c|}{56.09}      & 44.67  \\ \cline{2-18} 
                              & RR      & \multicolumn{1}{c|}{27.20}   & \multicolumn{1}{c|}{85.83}     & \multicolumn{1}{c|}{57.11}      & 45.91  & \multicolumn{1}{c|}{25.79}    & \multicolumn{1}{c|}{85.18}     & \multicolumn{1}{c|}{56.17}      & 45.20  & \multicolumn{1}{c|}{27.32}    & \multicolumn{1}{c|}{85.65}     & \multicolumn{1}{c|}{57.77}      & 45.87  & \multicolumn{1}{c|}{25.82}    & \multicolumn{1}{c|}{85.10}     & \multicolumn{1}{c|}{56.32}      & 45.96  \\ \hline
\end{tabular}}
\caption{Automatic evaluation scores on ExpertQA-Bio : we compare ROUGE-L, BERTScore, MoverScore, BLEURT on different settings. The superior scores among the same models are highlighted in bold.}
\label{TableR2}
\end{table}

\begin{figure*}
\centering
\subfloat[GPT-4 ]{\includegraphics[width = 0.5\textwidth]{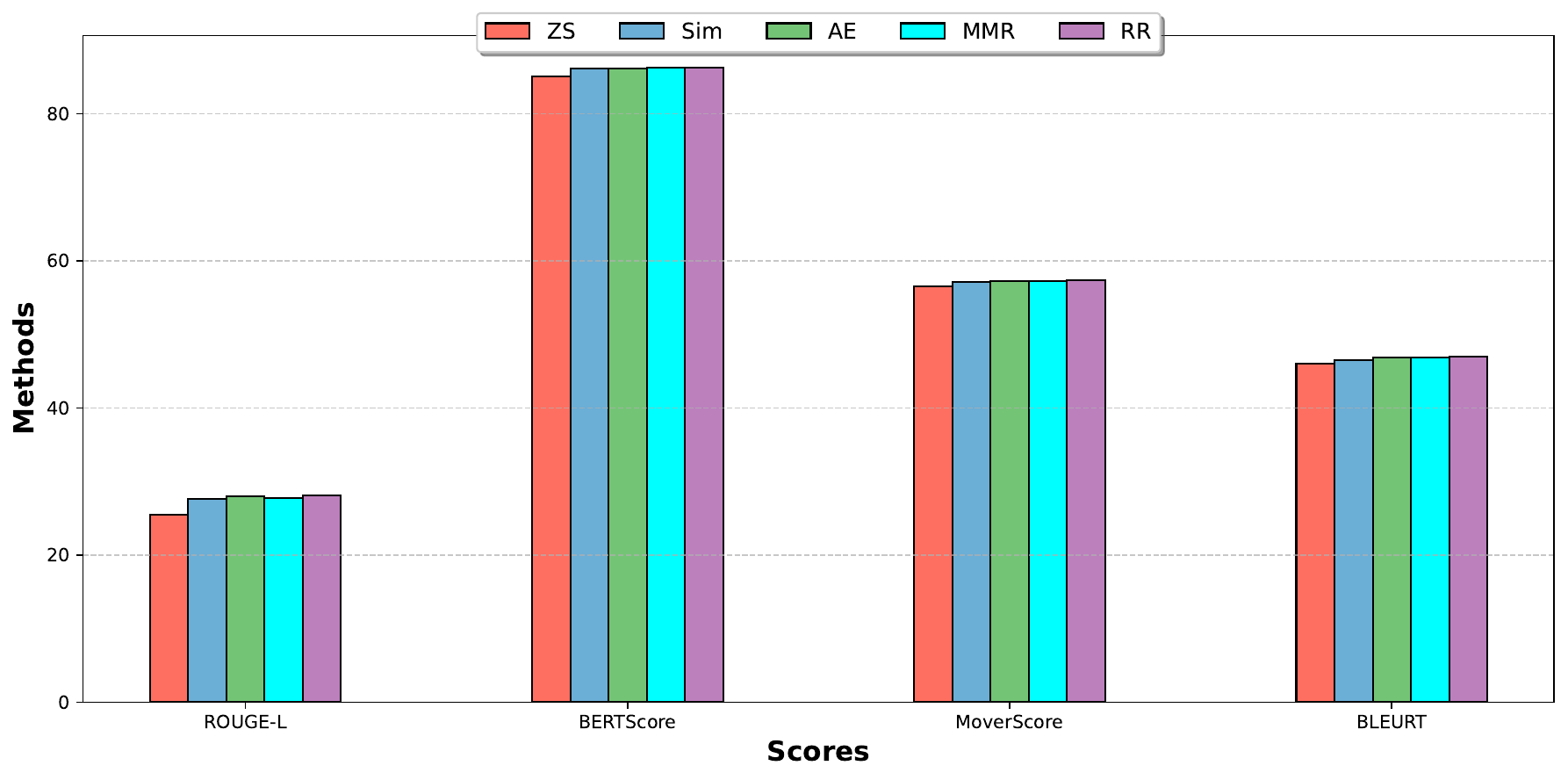}}
\subfloat[LLaMA2-13b]{\includegraphics[ width = 0.5\textwidth]{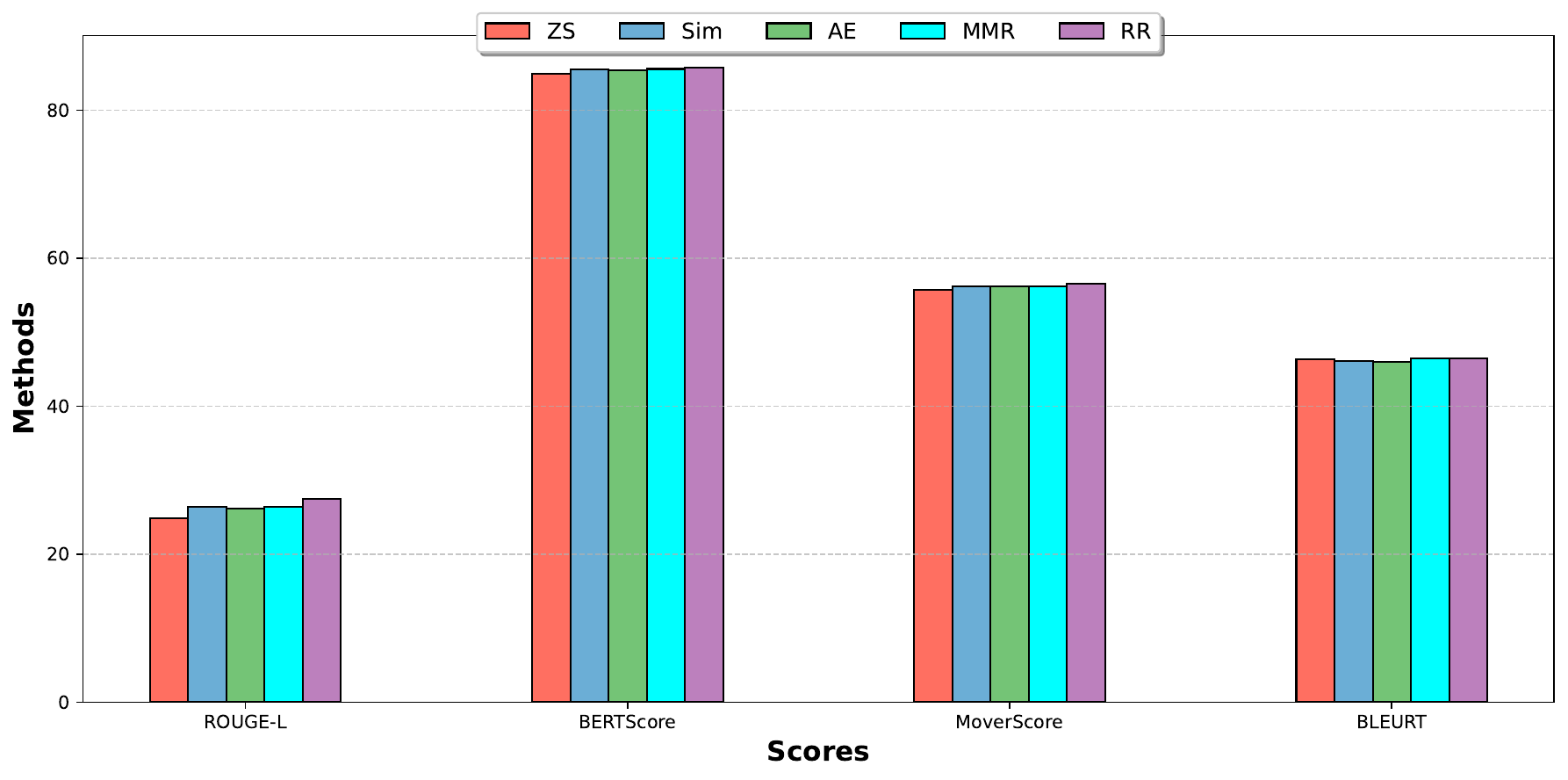}}\\
\subfloat[Galactica   30B]{\includegraphics[width = 0.5\textwidth]{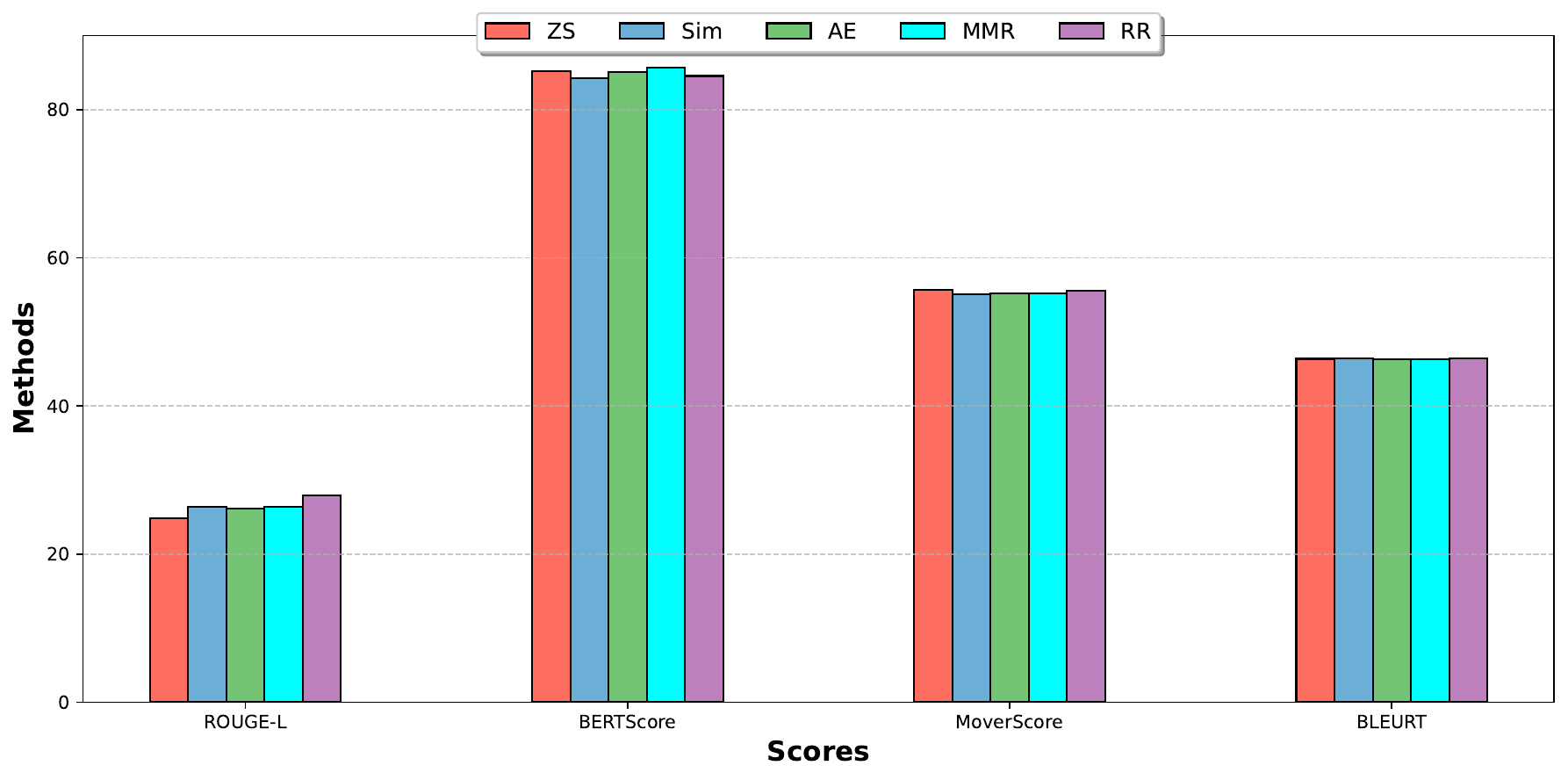}}
\subfloat[BioPars]{\includegraphics[width = 0.5\textwidth]{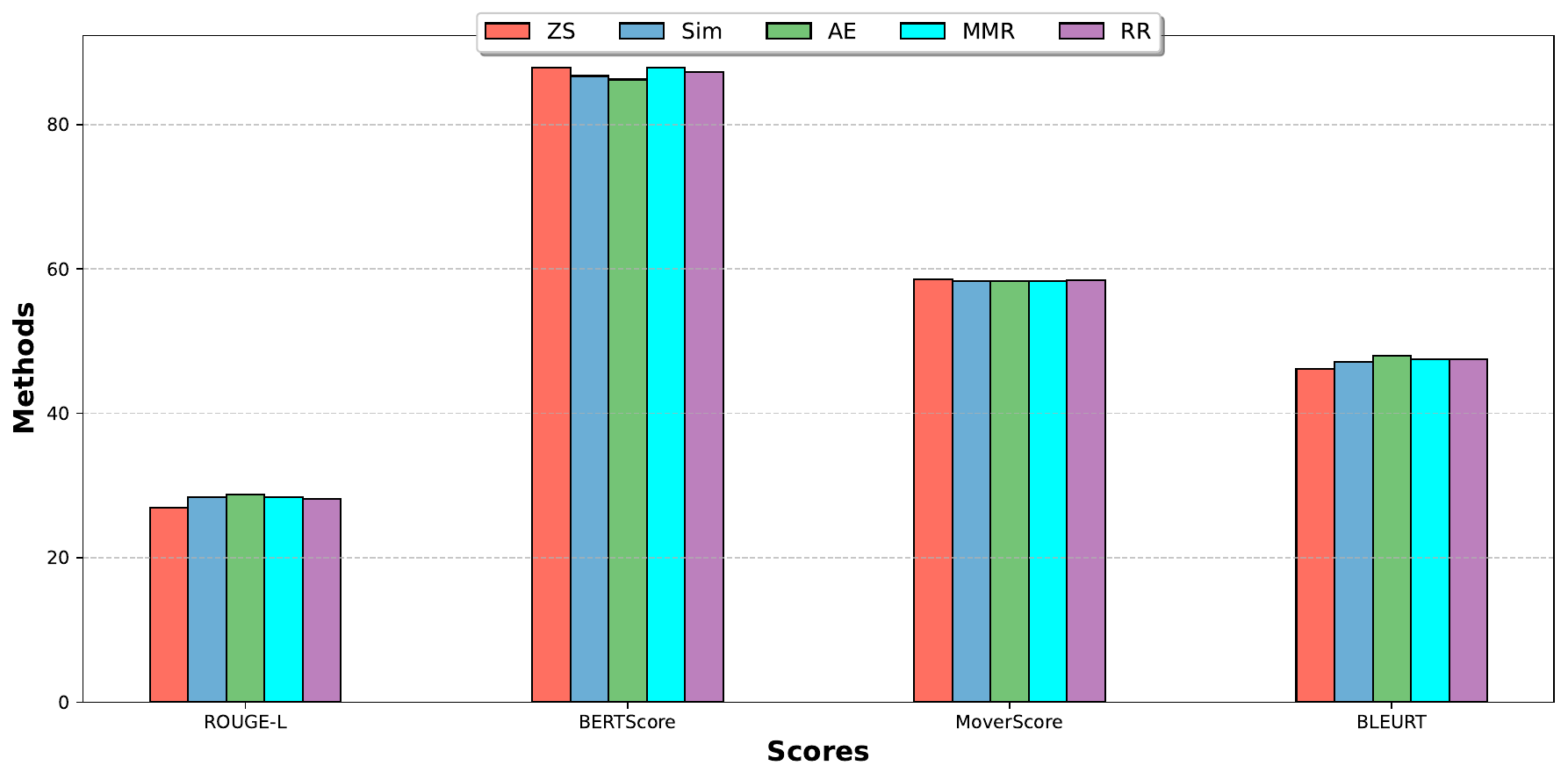}}\\
\caption{Comparative bar plot of different approaches on the  ExpertQA-Med.}
\label{BarChartQME}
\end{figure*}

The evaluation results of the proposed approach and the comparative approaches on the ExpertQA-Med dataset are given in Table \ref{TableR3}. The GPT-4 approach achieved successful results on this dataset compared to the two LLaMA2-13b and Galactica 30B models. This approach achieved a maximum ROUGE-L = 28.08, while the two LLaMA2-13b and Galactica 30B models achieved maximum ROUGE-L of 26.49 and 27.97, respectively. In terms of BERTScore, the GPT-4 model also recorded better results. The GPT-4 model also achieved a maximum MoverScore = 57.32 and BLEURT = 47.0. The BioPars model also achieved high results on this dataset. This model achieved ROUGE-L = 28.72 using the AE method, the highest value reported in all the tested models. The model also achieved a BERTScore of 87.92, a 1.62 improvement over the GPT-4 model. In the two evaluation metrics, MoverScore and BLEURT, the BioPars model achieved 58.54 and 47.95 at maximum, respectively.
The comparative bar plot of different approaches on the  ExpertQA-Med is shown in Figure \ref{BarChartQME}.
\begin{table}[]
\resizebox{\textwidth}{!}{%
\begin{tabular}{|l|l|cccc|cccc|cccc|cccc|}
\hline
Dataset                       & Methods & \multicolumn{4}{c|}{GPT-4}                                                                               & \multicolumn{4}{c|}{LLaMA2-13b}                                                                           & \multicolumn{4}{c|}{Galactica   30B}                                                                      & \multicolumn{4}{c|}{BioPars:}                                                                             \\ \hline
                              &         & \multicolumn{1}{c|}{ROUGE-L} & \multicolumn{1}{c|}{BERTScore} & \multicolumn{1}{c|}{MoverScore} & BLEURT & \multicolumn{1}{c|}{ROUGE- L} & \multicolumn{1}{c|}{BERTScore} & \multicolumn{1}{c|}{MoverScore} & BLEURT & \multicolumn{1}{c|}{ROUGE- L} & \multicolumn{1}{c|}{BERTScore} & \multicolumn{1}{c|}{MoverScore} & BLEURT & \multicolumn{1}{c|}{ROUGE- L} & \multicolumn{1}{c|}{BERTScore} & \multicolumn{1}{c|}{MoverScore} & BLEURT \\ \hline
\multirow{5}{*}{ExpertQA-Med} & ZS      & \multicolumn{1}{c|}{25.45}   & \multicolumn{1}{c|}{85.11}     & \multicolumn{1}{c|}{56.50}      & 45.98  & \multicolumn{1}{c|}{24.86}    & \multicolumn{1}{c|}{84.89}     & \multicolumn{1}{c|}{55.74}      & 46.32  & \multicolumn{1}{c|}{24.86}    & \multicolumn{1}{c|}{85.23}     & \multicolumn{1}{c|}{55.72}      & 46.36  & \multicolumn{1}{c|}{26.91}    & \multicolumn{1}{c|}{87.92}     & \multicolumn{1}{c|}{58.54}      & 46.12  \\ \cline{2-18} 
                              & Sim     & \multicolumn{1}{c|}{27.61}   & \multicolumn{1}{c|}{86.10}     & \multicolumn{1}{c|}{57.13}      & 46.47  & \multicolumn{1}{c|}{26.4}     & \multicolumn{1}{c|}{85.50}     & \multicolumn{1}{c|}{56.23}      & 46.15  & \multicolumn{1}{c|}{26.4}     & \multicolumn{1}{c|}{84.22}     & \multicolumn{1}{c|}{55.12}      & 46.45  & \multicolumn{1}{c|}{28.36}    & \multicolumn{1}{c|}{86.71}     & \multicolumn{1}{c|}{58.34}      & 47.12  \\ \cline{2-18} 
                              & AE      & \multicolumn{1}{c|}{27.98}   & \multicolumn{1}{c|}{86.12}     & \multicolumn{1}{c|}{57.25}      & 46.80  & \multicolumn{1}{c|}{26.15}    & \multicolumn{1}{c|}{85.36}     & \multicolumn{1}{c|}{56.17}      & 46.02  & \multicolumn{1}{c|}{26.15}    & \multicolumn{1}{c|}{85.12}     & \multicolumn{1}{c|}{55.17}      & 46.33  & \multicolumn{1}{c|}{28.72}    & \multicolumn{1}{c|}{86.23}     & \multicolumn{1}{c|}{58.27}      & 47.95  \\ \cline{2-18} 
                              & MMR     & \multicolumn{1}{c|}{27.78}   & \multicolumn{1}{c|}{86.22}     & \multicolumn{1}{c|}{57.28}      & 46.84  & \multicolumn{1}{c|}{26.42}    & \multicolumn{1}{c|}{85.57}     & \multicolumn{1}{c|}{56.24}      & 46.41  & \multicolumn{1}{c|}{26.43}    & \multicolumn{1}{c|}{85.73}     & \multicolumn{1}{c|}{55.24}      & 46.34  & \multicolumn{1}{c|}{28.44}    & \multicolumn{1}{c|}{87.91}     & \multicolumn{1}{c|}{58.26}      & 47.44  \\ \cline{2-18} 
                              & RR      & \multicolumn{1}{c|}{28.08}   & \multicolumn{1}{c|}{86.30}     & \multicolumn{1}{c|}{57.32}      & 47.00  & \multicolumn{1}{c|}{27.49}    & \multicolumn{1}{c|}{85.80}     & \multicolumn{1}{c|}{56.58}      & 46.47  & \multicolumn{1}{c|}{27.97}    & \multicolumn{1}{c|}{84.55}     & \multicolumn{1}{c|}{55.51}      & 46.41  & \multicolumn{1}{c|}{28.12}    & \multicolumn{1}{c|}{87.26}     & \multicolumn{1}{c|}{58.38}      & 47.43  \\ \hline
\end{tabular}}
\caption{Automatic evaluation scores on ExpertQA-Bio: we compare ROUGE-L, BERTScore, MoverScore, BLEURT on different settings. The superior scores among the same models are highlighted in bold.}
\label{TableR3}
\end{table}

\begin{figure*}
\centering
\subfloat[GPT-4 ]{\includegraphics[width = 0.5\textwidth]{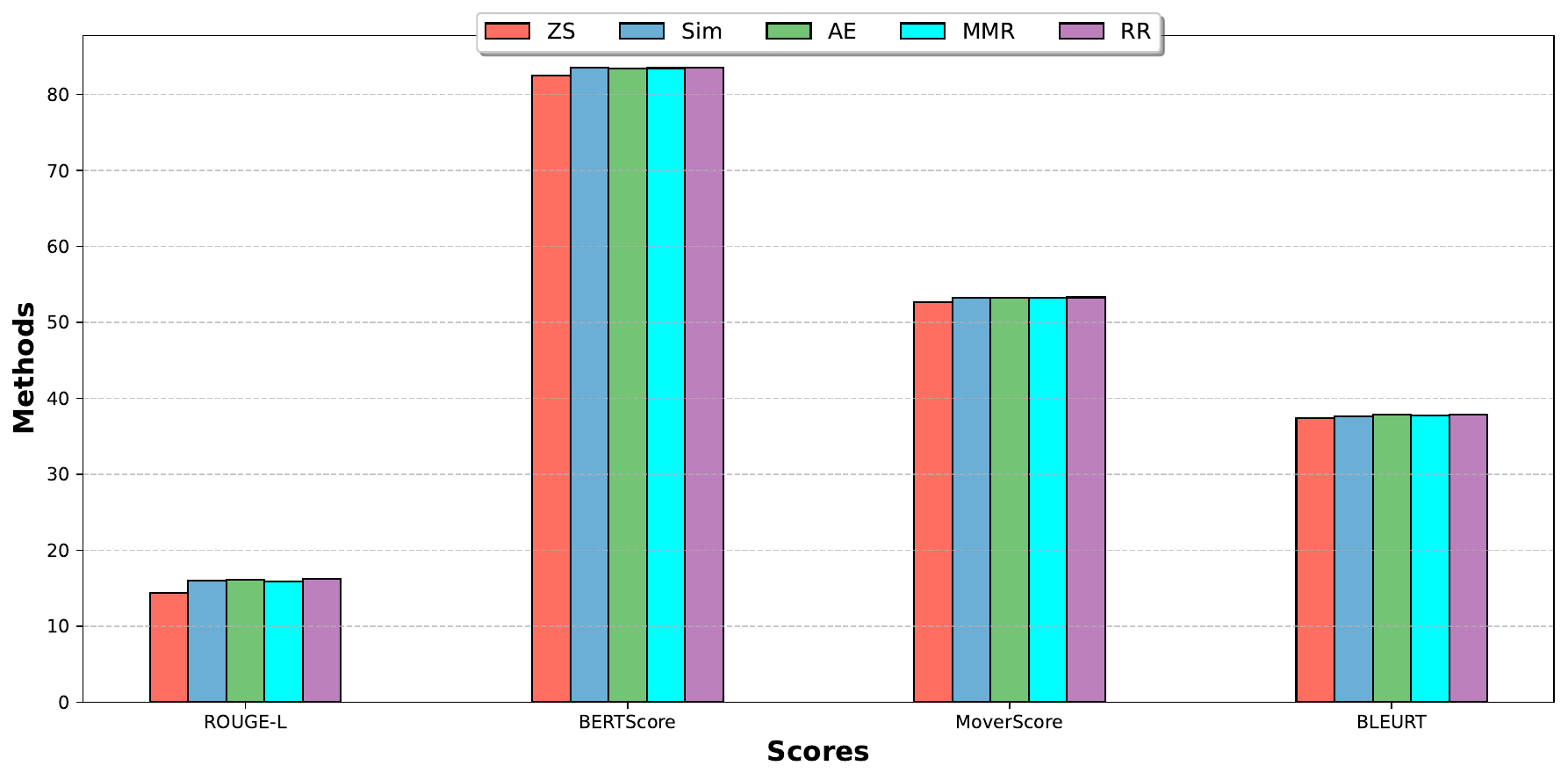}}
\subfloat[LLaMA2-13b]{\includegraphics[ width = 0.5\textwidth]{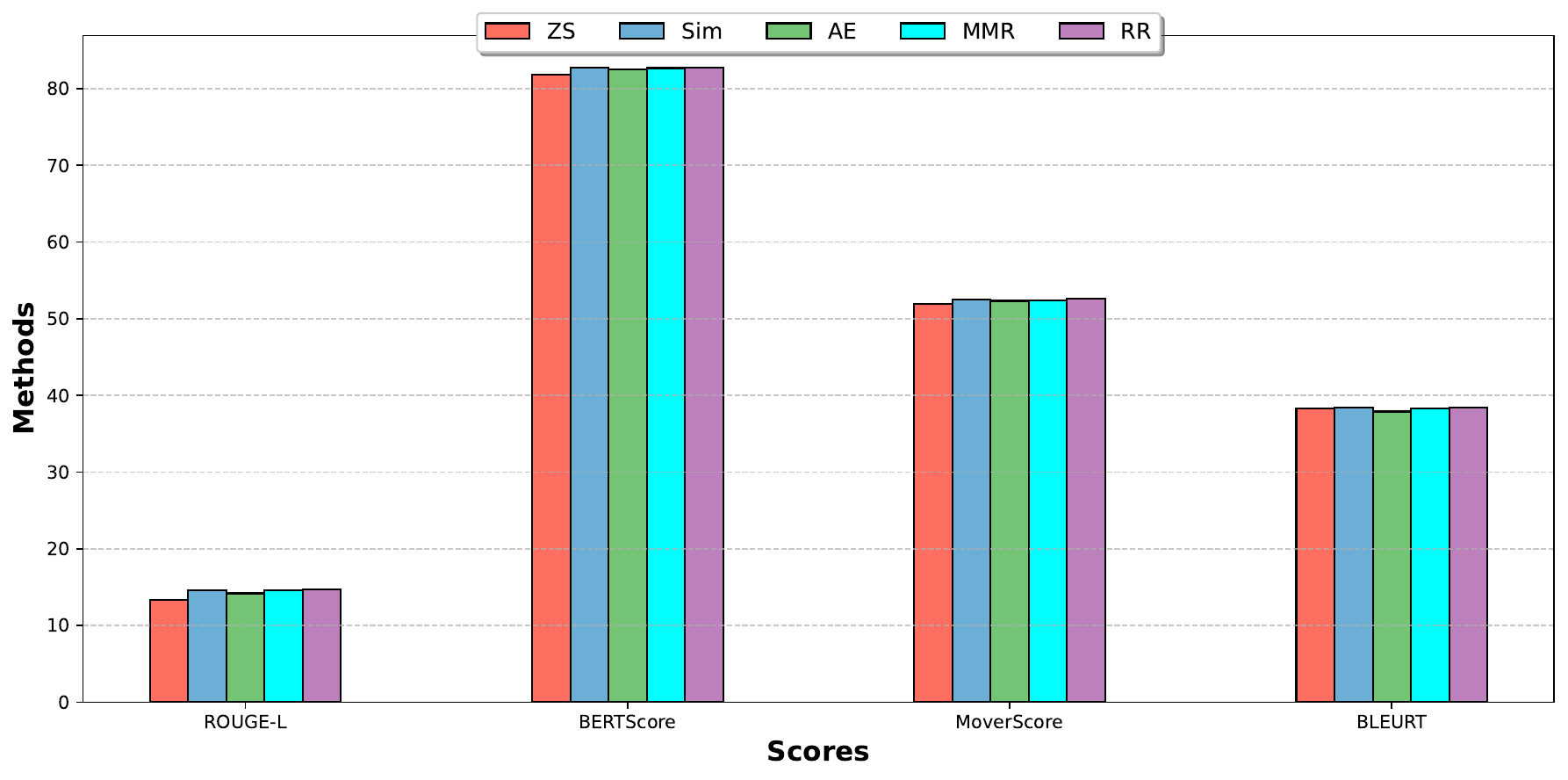}}\\
\subfloat[Galactica   30B]{\includegraphics[width = 0.5\textwidth]{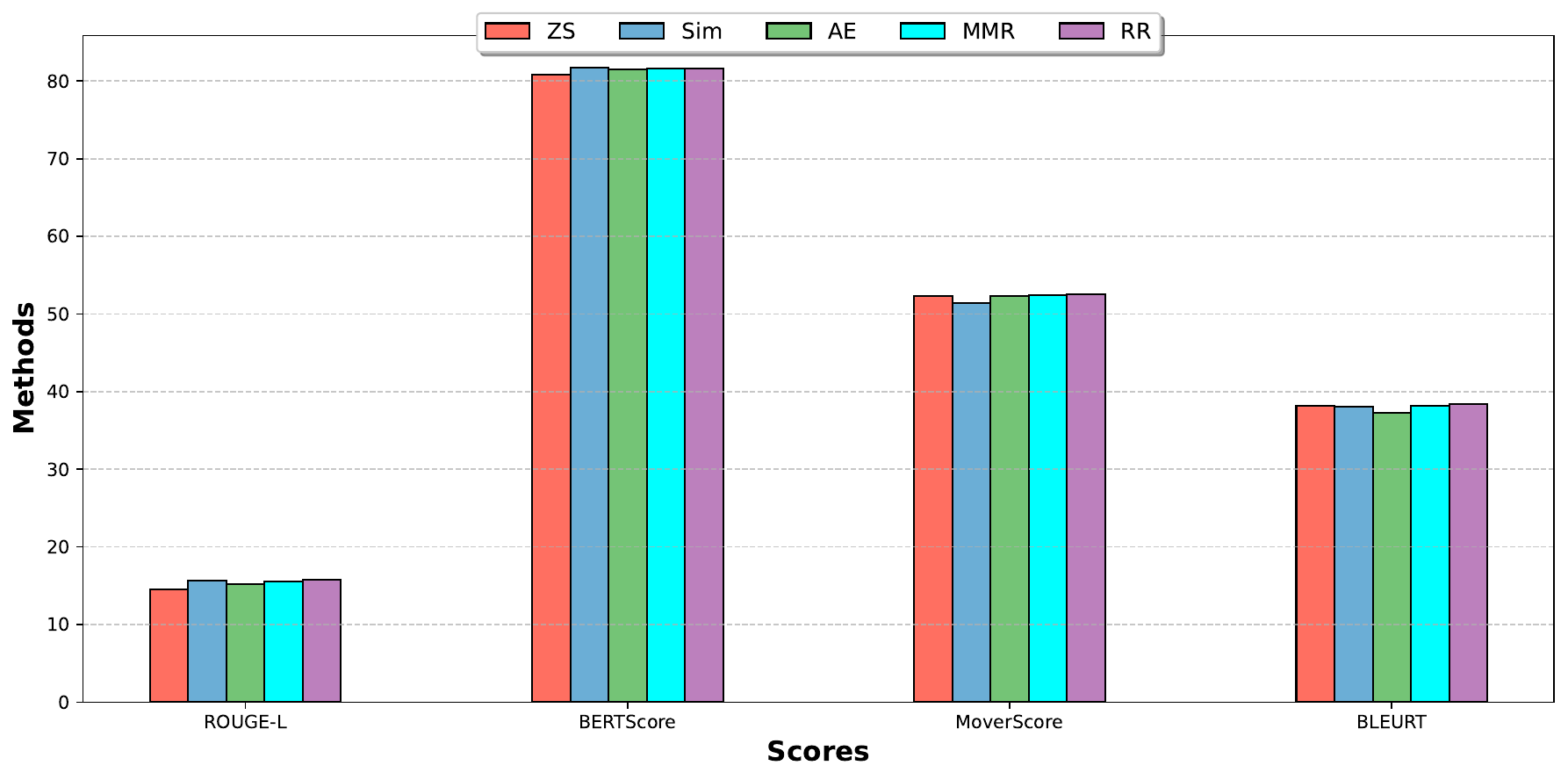}}
\subfloat[BioPars]{\includegraphics[width = 0.5\textwidth]{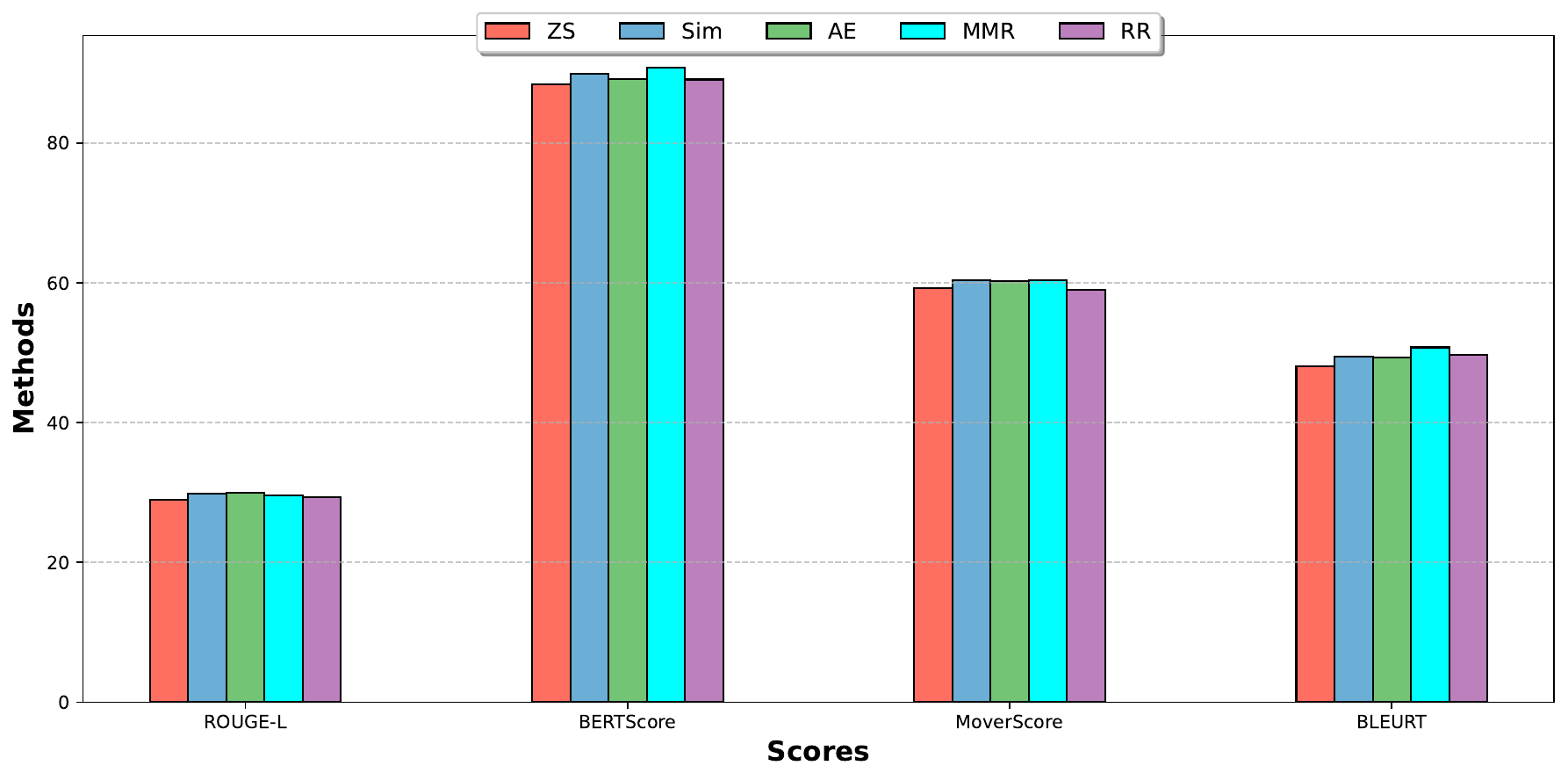}}\\
\caption{comparative bar plot of different approaches on the  MedicationQA.}
\label{BarChartMEdu}
\end{figure*}

The results of the different approaches and the proposed approach on the MedicationQA dataset are shown in Table \ref{TableR41}. The GPT-4 and LLaMA2-13b models achieved maximum ROUGE-L scores of 16.19 and 14.71, respectively. These two models also achieved maximum BERTScore scores of 83.59 and 82.79. The GPT-4 model performed better than LLaMA2-13b in the MoverScore evaluation criterion in the maximum case and worse than LLaMA2-13b in the BLEURT evaluation criterion in the maximum case. Galactica 30B performed worse than the other models in this dataset. This model achieved 15.73, 81.78, and 38.43 in the evaluation criteria. The BioPars model performed best among the tested models in this dataset. This model achieved ROUGE-L=18.14, which was a 1.19 improvement over GPT-4. This model achieved a maximum ERTScore=83.90 using the RR method. 54.32 and 39.87 were recorded in the maximum mode for MoverScore and BLEURT. The comparative bar plot of different approaches on the  MedicationQA is shown in Figure \ref{BarChartMEdu}.

\begin{table}[]
\resizebox{\textwidth}{!}{%
\begin{tabular}{|l|l|cccc|cccc|cccc|cccc|}
\hline
Dataset                       & Methods & \multicolumn{4}{c|}{GPT-4}                                                                               & \multicolumn{4}{c|}{LLaMA2-13b}                                                                           & \multicolumn{4}{c|}{Galactica   30B}                                                                      & \multicolumn{4}{c|}{BioPars}                                                                              \\ \hline
                              &         & \multicolumn{1}{c|}{ROUGE-L} & \multicolumn{1}{c|}{BERTScore} & \multicolumn{1}{c|}{MoverScore} & BLEURT & \multicolumn{1}{c|}{ROUGE- L} & \multicolumn{1}{c|}{BERTScore} & \multicolumn{1}{c|}{MoverScore} & BLEURT & \multicolumn{1}{c|}{ROUGE- L} & \multicolumn{1}{c|}{BERTScore} & \multicolumn{1}{c|}{MoverScore} & BLEURT & \multicolumn{1}{c|}{ROUGE- L} & \multicolumn{1}{c|}{BERTScore} & \multicolumn{1}{c|}{MoverScore} & BLEURT \\ \hline
\multirow{5}{*}{MedicationQA} & ZS      & \multicolumn{1}{c|}{14.41}   & \multicolumn{1}{c|}{82.55}     & \multicolumn{1}{c|}{52.62}      & 37.41  & \multicolumn{1}{c|}{13.30}    & \multicolumn{1}{c|}{81.81}     & \multicolumn{1}{c|}{51.96}      & 38.30  & \multicolumn{1}{c|}{14.54}    & \multicolumn{1}{c|}{80.82}     & \multicolumn{1}{c|}{52.33}      & 38.22  & \multicolumn{1}{c|}{17.23}    & \multicolumn{1}{c|}{82.54}     & \multicolumn{1}{c|}{53.56}      & 38.41  \\ \cline{2-18} 
                              & Sim     & \multicolumn{1}{c|}{16.05}   & \multicolumn{1}{c|}{83.56}     & \multicolumn{1}{c|}{53.23}      & 37.60  & \multicolumn{1}{c|}{14.60}    & \multicolumn{1}{c|}{82.73}     & \multicolumn{1}{c|}{52.47}      & 38.38  & \multicolumn{1}{c|}{15.64}    & \multicolumn{1}{c|}{81.78}     & \multicolumn{1}{c|}{51.44}      & 38.11  & \multicolumn{1}{c|}{18.14}    & \multicolumn{1}{c|}{83.54}     & \multicolumn{1}{c|}{54.26}      & 39.60  \\ \cline{2-18} 
                              & AE      & \multicolumn{1}{c|}{16.13}   & \multicolumn{1}{c|}{83.46}     & \multicolumn{1}{c|}{53.23}      & 37.87  & \multicolumn{1}{c|}{14.19}    & \multicolumn{1}{c|}{82.50}     & \multicolumn{1}{c|}{52.33}      & 37.90  & \multicolumn{1}{c|}{15.23}    & \multicolumn{1}{c|}{81.54}     & \multicolumn{1}{c|}{52.32}      & 37.31  & \multicolumn{1}{c|}{17.23}    & \multicolumn{1}{c|}{83.34}     & \multicolumn{1}{c|}{54.25}      & 39.87  \\ \cline{2-18} 
                              & MMR     & \multicolumn{1}{c|}{15.89}   & \multicolumn{1}{c|}{83.48}     & \multicolumn{1}{c|}{53.22}      & 37.73  & \multicolumn{1}{c|}{14.56}    & \multicolumn{1}{c|}{82.69}     & \multicolumn{1}{c|}{52.44}      & 38.31  & \multicolumn{1}{c|}{15.57}    & \multicolumn{1}{c|}{81.62}     & \multicolumn{1}{c|}{52.42}      & 38.22  & \multicolumn{1}{c|}{17.13}    & \multicolumn{1}{c|}{83.54}     & \multicolumn{1}{c|}{54.32}      & 39.73  \\ \cline{2-18} 
                              & RR      & \multicolumn{1}{c|}{16.19}   & \multicolumn{1}{c|}{83.59}     & \multicolumn{1}{c|}{53.30}      & 37.91  & \multicolumn{1}{c|}{14.71}    & \multicolumn{1}{c|}{82.79}     & \multicolumn{1}{c|}{52.59}      & 38.42  & \multicolumn{1}{c|}{15.73}    & \multicolumn{1}{c|}{81.64}     & \multicolumn{1}{c|}{52.53}      & 38.43  & \multicolumn{1}{c|}{17.22}    & \multicolumn{1}{c|}{83.9}      & \multicolumn{1}{c|}{54.21}      & 39.22  \\ \hline
\end{tabular}}
\caption{Automatic evaluation scores on MedicationQA: we compare ROUGE-L, BERTScore, MoverScore, BLEURT on different settings. The superior scores among the same models are highlighted in bold.}
\label{TableR41}
\end{table}

\begin{figure*}
\centering
\subfloat[GPT-4 ]{\includegraphics[width = 0.5\textwidth]{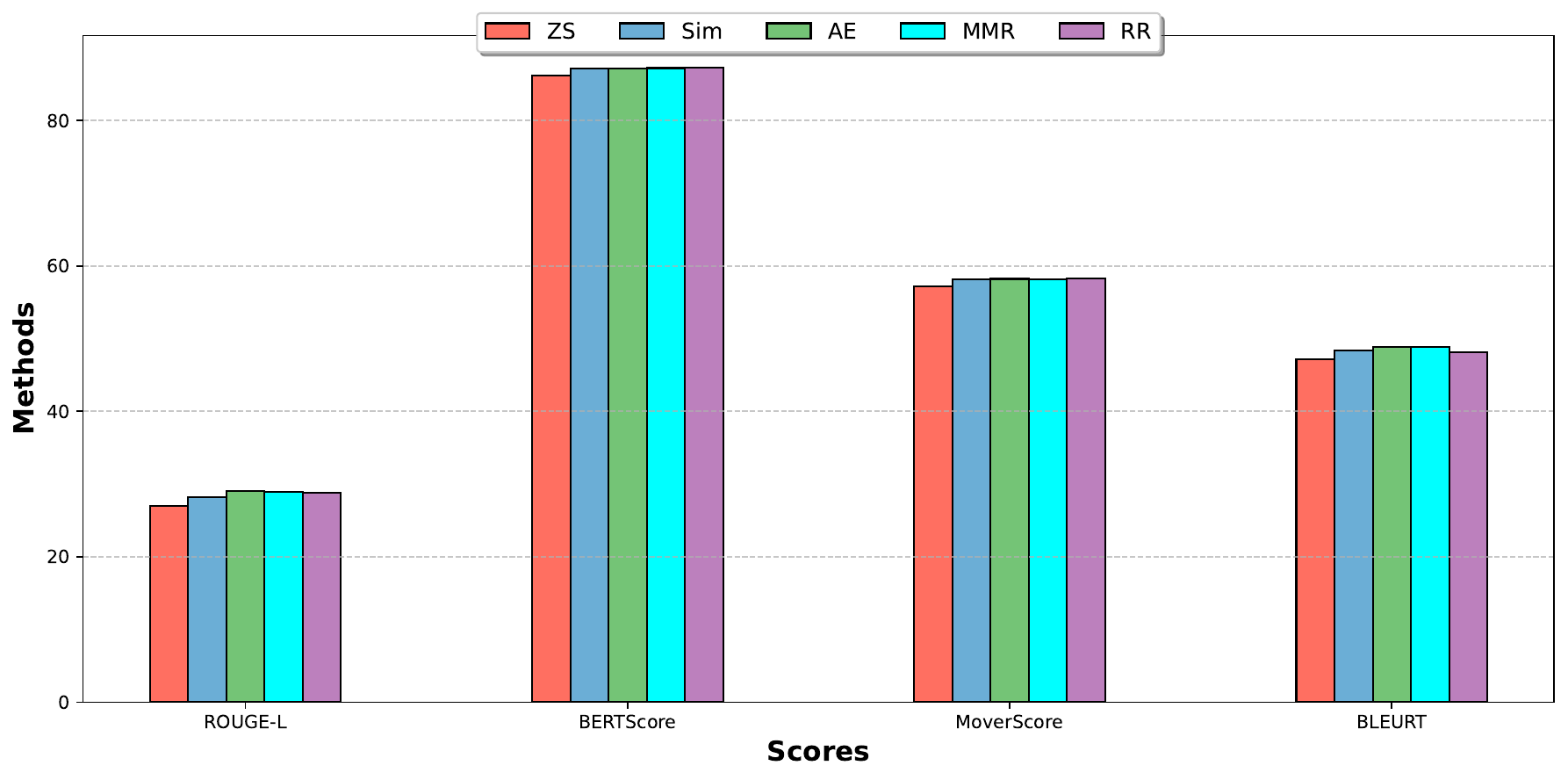}}
\subfloat[LLaMA2-13b]{\includegraphics[ width = 0.5\textwidth]{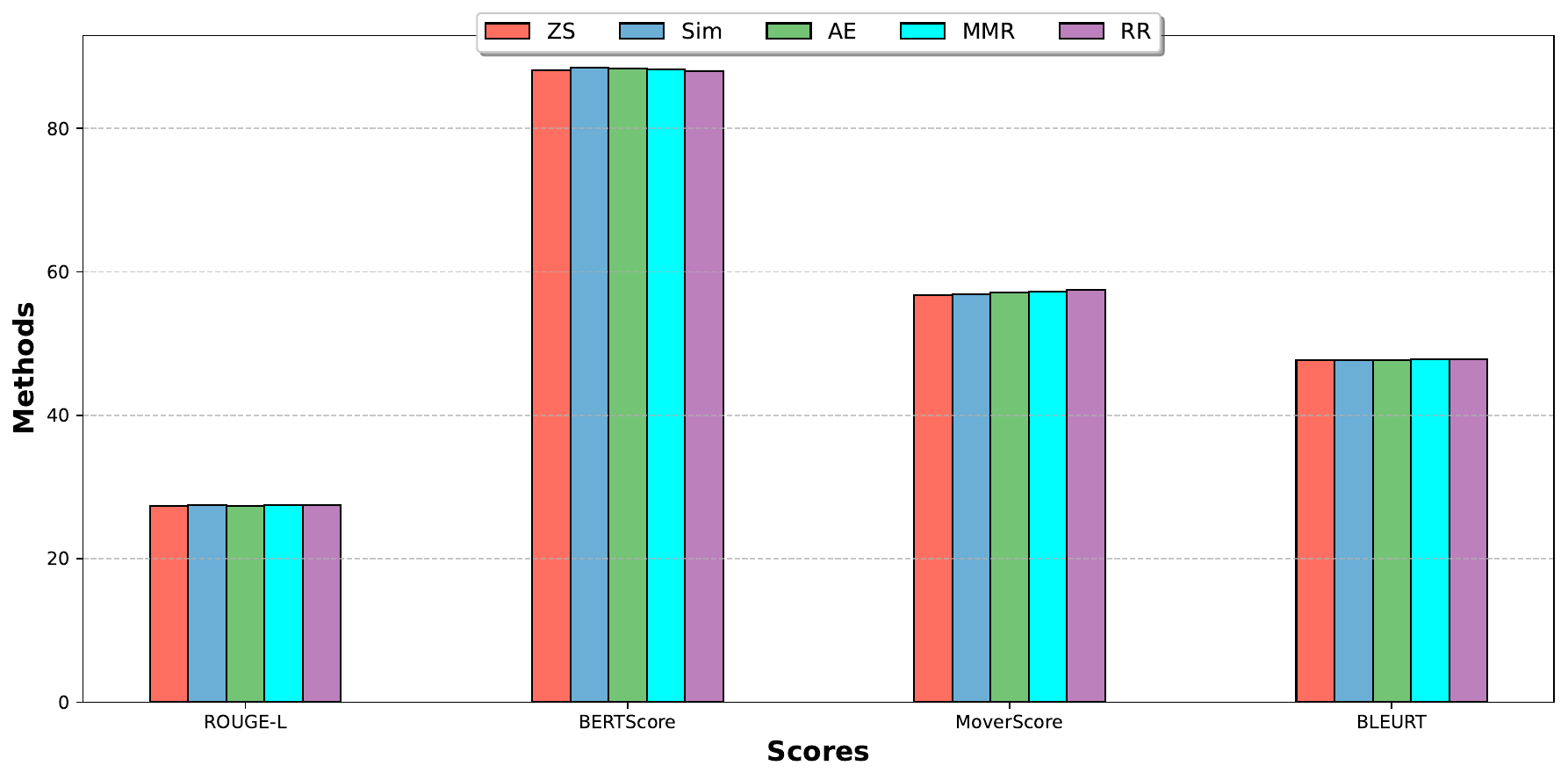}}\\
\subfloat[Galactica   30B]{\includegraphics[width = 0.5\textwidth]{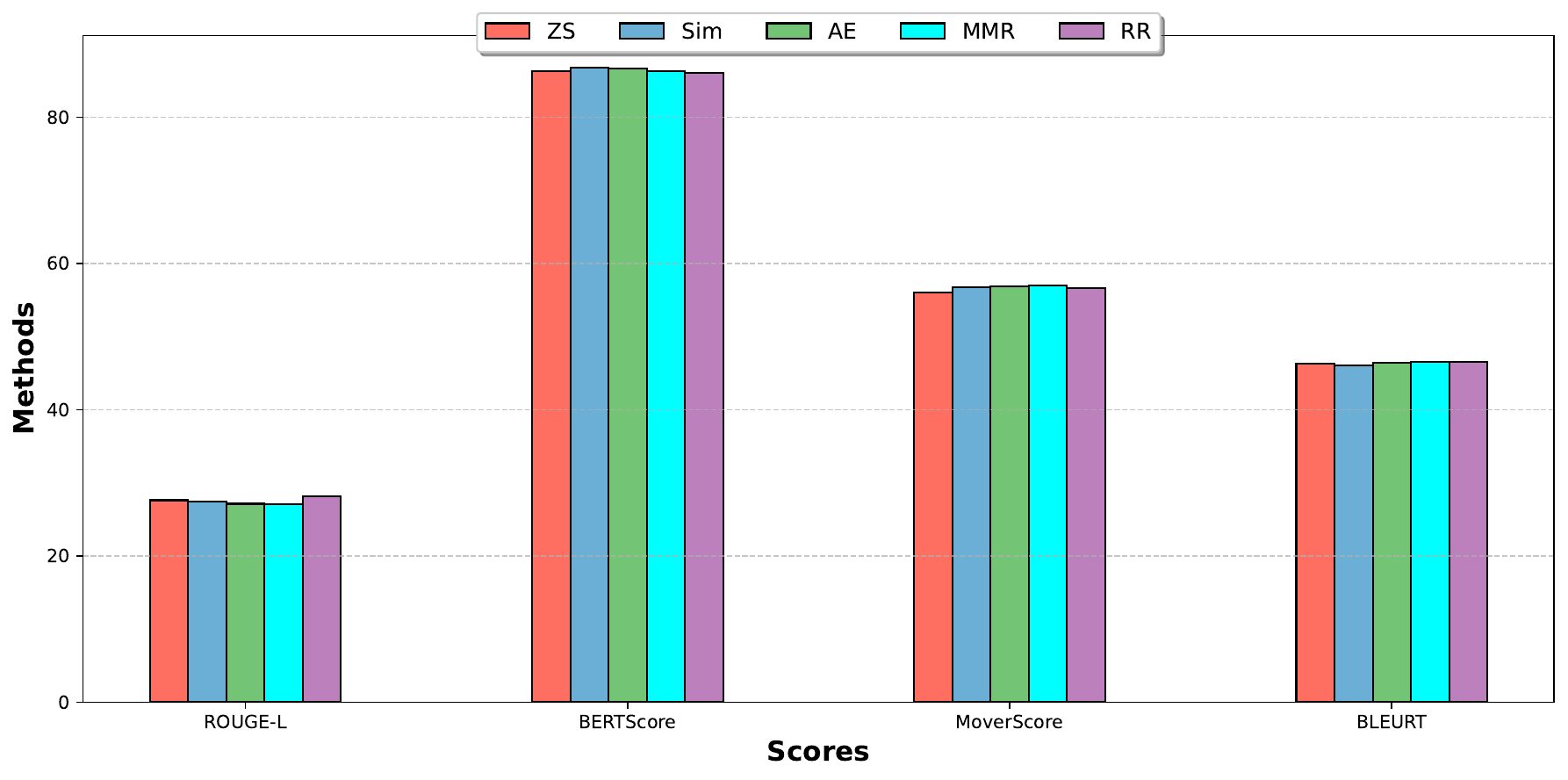}}
\subfloat[BioPars]{\includegraphics[width = 0.5\textwidth]{BioBioPPars.pdf}}\\
\caption{The comparative bar plot of different approaches on the  BioParsQA.}
\label{BarChartBioParsQA}
\end{figure*}

The evaluation results of the different approaches on the BioParsQA dataset are given in Table \ref{TableR4}. Initially, the GPT-4 approach achieved ROUGE-L=28.99 using the AE evaluation method. This approach also achieved maximum BERTScore=87.32, MoverScore=58.32, and BLEURT=48.82 using the RR, RR, and AE evaluation methods. The LLaMA2-13b approach performed worse than GPT-4 in the ROUGE-L evaluation criterion. This approach was able to achieve maximum ROUGE-L=88.52 using the Sim method. The BERTScore values in this approach are much higher than GPT-4, and all methods achieved BERTScore above 0.88. This approach's maximum MoverScore and BLEURT were 57.51 and 47.82, respectively. The maximum ROUGE-L in the Galactica 30B approach is 28.12. This approach also achieved maximum BERTscore, MoverScore, and BLEURT of 86.82, 56.94, and 46.56 on the BioParsQA dataset. The BioPars approach outperformed the other three models on BioParsQA. The model achieved a ROUGE-L score of 29.99, an improvement over GPT-4 1.0. The model achieved BERTscore=90.87 with the MMR method. The MoverScore and BLEURT values were also higher than the other three models. The model achieved MoverScore=60.43 and BLEURT=50.78. The comparative bar plot of different approaches on the  BioParsQA is shown in Figure \ref{BarChartBioParsQA}.

\begin{table}[]\resizebox{\textwidth}{!}{%
\begin{tabular}{|l|l|llll|llll|llll|llll|}
\hline
Dataset                    & Methods & \multicolumn{4}{c|}{GPT-4}                                                                               & \multicolumn{4}{c|}{LLaMA2-13b}                                                                           & \multicolumn{4}{c|}{Galactica   30B}                                                                      & \multicolumn{4}{c|}{BioPars}                                                                              \\ \hline
                           &         & \multicolumn{1}{l|}{ROUGE-L} & \multicolumn{1}{l|}{BERTScore} & \multicolumn{1}{l|}{MoverScore} & BLEURT & \multicolumn{1}{l|}{ROUGE- L} & \multicolumn{1}{l|}{BERTScore} & \multicolumn{1}{l|}{MoverScore} & BLEURT & \multicolumn{1}{l|}{ROUGE- L} & \multicolumn{1}{l|}{BERTScore} & \multicolumn{1}{l|}{MoverScore} & BLEURT & \multicolumn{1}{l|}{ROUGE- L} & \multicolumn{1}{l|}{BERTScore} & \multicolumn{1}{l|}{MoverScore} & BLEURT \\ \hline
\multirow{5}{*}{BioParsQA} & ZS      & \multicolumn{1}{l|}{27.01}   & \multicolumn{1}{l|}{86.13}     & \multicolumn{1}{l|}{57.23}      & 47.12  & \multicolumn{1}{l|}{27.32}    & \multicolumn{1}{l|}{88.12}     & \multicolumn{1}{l|}{56.78}      & 47.72  & \multicolumn{1}{l|}{27.61}    & \multicolumn{1}{l|}{86.27}     & \multicolumn{1}{l|}{56.01}      & 46.32  & \multicolumn{1}{l|}{28.95}    & \multicolumn{1}{l|}{88.45}     & \multicolumn{1}{l|}{59.23}      & 48.11  \\ \cline{2-18} 
                           & Sim     & \multicolumn{1}{l|}{28.21}   & \multicolumn{1}{l|}{87.15}     & \multicolumn{1}{l|}{58.12}      & 48.33  & \multicolumn{1}{l|}{27.50}    & \multicolumn{1}{l|}{88.52}     & \multicolumn{1}{l|}{56.90}      & 47.71  & \multicolumn{1}{l|}{27.40}    & \multicolumn{1}{l|}{86.82}     & \multicolumn{1}{l|}{56.74}      & 46.00  & \multicolumn{1}{l|}{29.84}    & \multicolumn{1}{l|}{89.91}     & \multicolumn{1}{l|}{60.43}      & 49.42  \\ \cline{2-18} 
                           & AE      & \multicolumn{1}{l|}{28.99}   & \multicolumn{1}{l|}{87.13}     & \multicolumn{1}{l|}{58.22}      & 48.82  & \multicolumn{1}{l|}{27.34}    & \multicolumn{1}{l|}{88.31}     & \multicolumn{1}{l|}{57.11}      & 47.72  & \multicolumn{1}{l|}{27.13}    & \multicolumn{1}{l|}{86.69}     & \multicolumn{1}{l|}{56.85}      & 46.44  & \multicolumn{1}{l|}{29.99}    & \multicolumn{1}{l|}{89.23}     & \multicolumn{1}{l|}{60.23}      & 49.34  \\ \cline{2-18} 
                           & MMR     & \multicolumn{1}{l|}{28.91}   & \multicolumn{1}{l|}{87.21}     & \multicolumn{1}{l|}{58.21}      & 48.80  & \multicolumn{1}{l|}{27.46}    & \multicolumn{1}{l|}{88.19}     & \multicolumn{1}{l|}{57.22}      & 47.82  & \multicolumn{1}{l|}{27.11}    & \multicolumn{1}{l|}{86.32}     & \multicolumn{1}{l|}{56.94}      & 46.54  & \multicolumn{1}{l|}{29.54}    & \multicolumn{1}{l|}{90.87}     & \multicolumn{1}{l|}{60.43}      & 50.78  \\ \cline{2-18} 
                           & RR      & \multicolumn{1}{l|}{28.75}   & \multicolumn{1}{l|}{87.32}     & \multicolumn{1}{l|}{58.32}      & 48.09  & \multicolumn{1}{l|}{27.49}    & \multicolumn{1}{l|}{88.01}     & \multicolumn{1}{l|}{57.51}      & 47.80  & \multicolumn{1}{l|}{28.12}    & \multicolumn{1}{l|}{86.01}     & \multicolumn{1}{l|}{56.65}      & 46.56  & \multicolumn{1}{l|}{29.34}    & \multicolumn{1}{l|}{89.12}     & \multicolumn{1}{l|}{59.01}      & 49.66  \\ \hline
\end{tabular}}
\caption{Automatic evaluation scores on BioParsQA: we compare ROUGE-L, BERTScore, MoverScore, BLEURT on different settings. The superior scores among the same models are highlighted in bold.}
\label{TableR4}
\end{table}

\section{Conclusion and Limitations}
We introduce BioPars, the first comprehensive Persian medical LLM that implements the entire training pipeline from pre-training to final testing, outperforming other open-source medical LLMs. Further experiments highlight the importance of pre-training this model for the Persian medical domain. We also collected a large set of Persian medical data to train the LLM model from Persian medical references, online sites, textbooks, and scientific articles. We also created a large-scale Persian medical conversation dataset, BioParsQA. Despite these achievements, we recognize the limitations of the model. BioPars cannot guarantee the accuracy of answers in all its aspects. Given the serious consequences of incorrect data in the medical domain, we strongly suggest that users exercise caution and seek expert advice when dealing with the generated information.

On the other hand, any incorrect medical suggestion and decision may lead to serious consequences. In the future, we will focus on increasing data reliability, integrating more real-time data with monitoring, and incorporating non-textual multimodal information for a more comprehensive and accurate medical service. BioPars is primarily a research model and cannot be considered a substitute for professional medical advice. This model serves as a milestone for LLMs in the Persian medical field.

\section*{Code and Data Availability}
The source code, pretrained weights, and evaluation scripts for the BioPars model are publicly available at the following GitHub repository: \url{https://github.com/amirap80/BioPars}.

\section*{Authors Contribution Statement}
All authors contributed equally to data collection, analysis, and drafting of the manuscript. The corresponding author reviewed and approved the final version of the manuscript.

\section*{Funding}
This research received no funding.

\section*{Ethics Declarations}
This study does not involve identifiable human data, human participants, or questionnaire-based surveys; all datasets consist of publicly available textual resources, and therefore, no institutional ethics approval was required.

\end{document}